\documentclass[10pt,twocolumn,letterpaper]{article}

\usepackage{cvpr}
\usepackage{times}
\usepackage{epsfig}
\usepackage{graphicx}
\usepackage{amsmath}
\usepackage{amssymb}
\usepackage{booktabs}
\usepackage{float}
\usepackage{pifont}
\newcommand{\cmark}{\textcolor{green}{\ding{51}}}%
\newcommand{\xmark}{\textcolor{red}{\ding{55}}}%

\cvprfinalcopy 

\usepackage{multirow}
\usepackage[pagebackref,breaklinks,colorlinks]{hyperref}
\hypersetup{citecolor={orange}, linkcolor={blue}}
\usepackage[symbol]{footmisc}
\usepackage{orcidlink}

\usepackage[capitalize]{cleveref}
\crefname{section}{Sec.}{Secs.}
\Crefname{section}{Section}{Sections}
\Crefname{table}{Table}{Tables}
\crefname{table}{Tab.}{Tabs.}

\def\cvprPaperID{9146} 

\author{Karpikova Polina\orcidlink{0009-0009-3087-6613}*$^{1,2}$, Radionova Ekaterina\orcidlink{0009-0007-8051-5811}*$^1$, Yaschenko Anastasia\orcidlink{0009-0007-7942-5915}*$^{1,2}$, Spiridonov Andrei\orcidlink{0000-0001-7182-4486}*$^1$ \\
Kostyushko Leonid\orcidlink{0009-0006-6480-558X}$^3$, Fabbricatore Riccardo\orcidlink{0000-0002-6298-2885}$^1$, Ivakhnenko Aleksei\orcidlink{0000-0003-0555-3598}$^{1\dagger}$ 
\\
$^1$Samsung Research \quad
$^2$Higher School of Economics - Moscow, Russia
\\
$^3$Lomonosov Moscow State University - Russia
}

\begin{document}

\title{FIANCEE: Faster Inference of Adversarial Networks via Conditional Early Exits}

\maketitle
\footnotetext[1]{These authors contributed equally to this work}
\footnotetext[2]{ivakhnenko.aleksei@gmail.com}

\begin{abstract}
Generative DNNs are a powerful tool for image synthesis, but they are limited by their computational load. 
On the other hand, given a trained model and a task, \eg faces generation within a range of characteristics, the output image quality will be unevenly distributed among images with different characteristics. It follows, that we might restrain the model's complexity on some instances, maintaining a high quality.
We propose a method for diminishing computations by adding so-called early exit branches to the original architecture, and dynamically switching the computational path depending on how difficult it will be to render the output. We apply our method on two different SOTA models performing generative tasks: generation from a semantic map, and cross-reenactment of face expressions; showing it is able to output images with custom lower-quality thresholds. For a threshold of LPIPS $\leq0.1$, we diminish their computations by up to a half. This is especially relevant for real-time applications such as synthesis of faces, when quality loss needs to be contained, but most of the inputs need fewer computations than the complex instances.

\end{abstract}

\section{Introduction}
\label{sec:intro}

Image synthesis by generative adversarial networks (GANs) received great attention in the last years~\cite{wang2020ganreview,shamsolmoali2021synthreview}, its applications span from image-to-image translation~\cite{Isola2017ImagetoImageTW} to text-to-image rendering~\cite{frolov2021adv_text2img_review}, neural head avatars generation \cite{Drobyshev22megaportraits} and many more. However, this approach suffers from heavy computational burdens when challenged with producing photo-realistic images. Our work stems from the observation that deep neural networks (DNNs) output images with different but consistent quality when conditioned on certain parameters. Since their expressivity is uneven within the set of possibly generated images, it follows that for some examples, a simpler DNN may suffice in generating an output with the required quality.

On the other hand, approaches aimed at easing the heavy computational load of DNNs have been applied with great results, significantly decreasing redundant computations~\cite{alquahtani2021review, deng2020review}. While strategies such as pruning~\cite{Lecun1989sparse, molchanov2019importance, Shen_2022_CVPR} or knowledge distillation~\cite{caruana2006model, gou2021knowledge, cho2019efficacy} generate a DNN with fewer parameters, early exit (EE)~\cite{kaya2019shallowdeep, xing2020early} is a setup that allows for dynamic variation of the computational burden, and therefore presents itself as an ideal candidate for an image generation strategy aimed at outputting pictures of consistent quality, while avoiding excessive computation due to their irregular rendering difficulty.

Despite this, implementing EE strategies has remained out of the scope of studies on generative models. This is perhaps due to the fact that EE processes logits of intermediate layers, thus restricting their field of application to tasks where the latter are meaningful (\eg in classification), while excluding pipelines in which a meaningful output is given only at the last layer (\eg generative convolutional networks).

We propose a method that employs an EE strategy for image synthesis, dynamically routing the computational flow towards the needed exit in accordance to pictures' complexity, therefore reducing computational redundancy while maintaining consistent quality. To accomplish this, we employ three main elements, which constitute the novel contributions of our work.

First, we attach \textit{exit branches} to the original DNN (referred as the backbone), as portrayed in \cref{fig:pipeline}. These branches are built of lightweight version of the modules constituting the backbone architecture, their complexity can be tuned in accordance with the desired quality-cost relation. Their depth (\ie number of modules) varies in accordance to the number of backbone modules left after the point they get attached to. In this way, intermediate backbone logits are fairly processed.

In second place, we make use of a small \textit{database} of features, from which guiding examples are selected and used to condition image generation by concatenating them to the input of each branch. These features are obtained by processing a selection of images by the first layers of the backbone. Its presence yields a quality gain for earlier exits, at the expense of a small amount of memory and computations, thus harmonizing exits' output quality. This is extremely handy for settings where real-time rendering is needed and guiding examples can be readily provided, such as neural avatar generation.

Lastly, the third component of our workflow is a \textit{predictor}, namely a DNN trained on the outputs of our branches, and capable of indicating the exit needed for outputting an image of a given quality. This element is fundamental for ensuring a consistent lower-quality threshold, as we will see.

Our method is applicable to already trained models, but requires additional training for the newly introduced components.
We report its application to two distinct tasks of the image synthesis family, namely generation from a semantic map, and cross-reenactment of face expressions. Our main result may be summarized in this way: the method is easily applicable to already existing and trained generative models, it is capable of outputting images with custom lower-quality threshold by routing easier images to shorter computational paths, and the mean gain in terms of saved computations per quality loss is, respectively, $1.2\times10^3$, and $1.3\times10^3$ GFLOPs/LPIPS for the two applications. 

\section{Related work}
\label{sec:related work}

\subsection{Conditional generative adversarial networks}
Generative adversarial networks (GANs) are a class of generative frameworks based on the competition between two neural networks, namely a \textit{generator} and a \textit{discriminator}~\cite{goodfellow2020gan, goodfellow2014gan, gonog2019gan_review}. While the latter performs a classification task (decides whether a generated image is real or not), the former synthesises an image from a target distribution.

Conditional GANs are a variation of the original framework~\cite{mirza2014cgan}. Their architecture allows for the input of additional information, which is used to restrict the target space according to it. In this way, the network may be conditioned, for instance, by mask ~\cite{Isola2017ImagetoImageTW}, label~\cite{odena2017cgan}, or text~\cite{Reed2016gan_text-to-im}.

\subsection{Neural head avatars}
Recent years have seen the rise of neural head avatars as a practical method for creating head models. They allow to reenact a face with given expression and pose. Such models could be divided into two groups – the ones with latent geometry \cite{Burkov2020neural_head, Drobyshev22megaportraits, lempitsky2020bilayer, lempitsky2019fewshots, doukas2022talking} and those with 3d prior, e.g. head mesh  \cite{Doukas2021headgan, Khakhulin2022ROME, zheng2022implicit,Khakhulin2022ROME, grassal2022monocular, tianye2017flame, feng2021deca}. Additionally, there is a set of papers, targeting the whole human body, including the head and face, which could be divided by input data requirements. Some of them take only few images \cite{alldieck2022phorum}, others require a video \cite{kapper2021monocular, zheng2022slrf, fruhstuck2022insetgan, chen2021animatable, he2021arch++, kappel2021fidelity, yoon2022dynamic_humans, peng2022animatable_implicit, hu2021hvtr}. In this work, we refer to \cite{Drobyshev22megaportraits}, as the state-of-the-art method for one-shot, high-resolution neural head reenactment.

\subsection{Early exits}
Early exits are a computational-saving strategy employed mainly in classification tasks~\cite{scardapane2020should, laskaridis2021EEreview}. They are characterized by the addition of outputs to the DNN, from which an approximation of the final result can be obtained at a lower computational cost. They were rediscovered through the years as a standalone approach, despite being natively implemented in architectures such as Inception~\cite{szegedy2015CVPR} as a countermeasure to overfitting. Seldomly this approach has also been called cascade learning~\cite{marquez2018cascade, wang2017idk,leroux2017cascading}, adaptive neural network~\cite{bolukbasi2017adaptive} or simply branching~\cite{scardapane2020branching}. Proposed implementations differ on three design choices: exits' architecture, \ie what type of layers to use for processing the backbone's logits; where to append exits in order to spread evenly computations among them; and how to choose the computational path. The latter issue is often solved by implementing a confidence mechanism and selecting a single exit~\cite{zhou2020bert, teerapittayanon2016branchynet, kaya2019shallowdeep} or reusing predictions for further computations~\cite{wolczyk2021zero, xing2020early}. To a lesser extent, learnable exit policies have been proposed as well~\cite{chen2020learning_EE, dai2020learning_EE, scardapane2020branching}. 

\subsection{Predictor}
Changing computational path on a per-input basis has been proposed as a way for efficiently utilizing a single exit during inference~\cite{odena2017rl_pred, li2022pred_eng}. Our approach is inspired by a technique pioneered in the field of neural architecture search: the use of a so-called \textit{predictor} to speed up the performance estimation of a given architecture~\cite{cai2020once, wen2020neural_pred}, as well as in natural language processing \cite{xin2021berxit, elbayad2020depth}, and has been applied to inference through early exits for resource-constrained edge AI. \cite{dong2022prediction}.

\begin{figure*}
    \centering
    \includegraphics[width=0.8\textwidth]{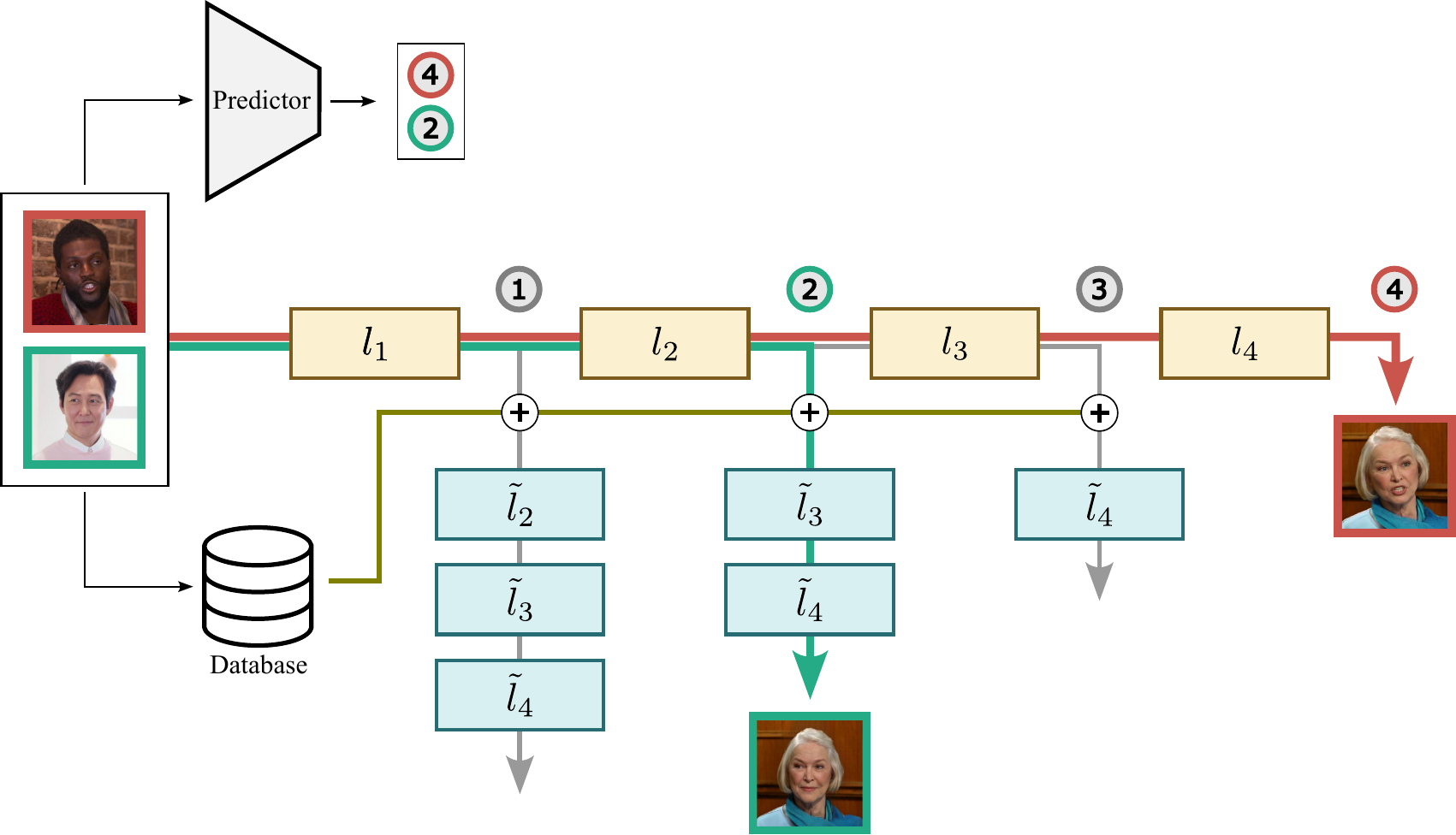}
    \caption{Our pipeline. In this example, the backbone generator is composed of yellow modules $l_1$ through $l_4$. We append three branches, thus adding early exits $1$ through $3$. Each branch has a different depth, and is composed of lightweight modules $\tilde{l}_i$. We show the computational path for two distinct inputs. The top input (red contour) is fed to the predictor (shown at the top), which deems it complex enough to require exit $4$ for the given quality threshold. The bottom input (green contour), instead, needs only exit $2$ to satisfy quality requirements. For both examples, an auxiliary image is retrieved from the database, in order to guide the synthesis.}
    \label{fig:pipeline}
\end{figure*}

\subsection{Database use}
Early image synthesis methods were based on the retrieval of examples from large image datasets~\cite{hays2007millions, isola2013SceneCollaging,johnson2009cg2real, lalonde2007photoclipart}. This is in contrast with contemporary DNN techniques, which rely on a large number of parameters to output photorealistic images. On the other hand, semi-parametric generation has been proposed in order to exploit strengths of both approaches~\cite{tsai2017harmony, qi2018semiparametric, Shi2022retriveral}. In particular, the use of patches, reminiscent of the old methods, seems to achieve great accuracy~\cite{granot2022drop_the_gan, li2019semi-param_from_graph, tseng2020patch_retrieval}.

Storing a large image database poses a problem when it comes to querying it in order to extract the needed sample. Looking for guiding images, we must employ an algorithm that will quickly find a similar picture or patch. To this end, we borrow from the literature that employs caches~\cite{grave2017improving, orhan2018simple_cache} and in particular nearest-neighbours search~\cite{jia2021rethinking, granot2022drop_the_gan, zhang2021tip}, where pretrained models are used as visual feature extractors, and the weights of the image encoders are fixed.

\section{Our method}
\label{or approach}

Our goal is to implement an early exit mechanism into a GAN model in order to render quickly easier images. To this end, we implement three elements:

\subsection{Depth-varying exit branches}
\label{subsec:depth_varying_exits}
As discussed, GANs are composed of two competing DNNs: a generator $G$ and a discriminator $D$. The former is designed to synthesize arbitrary images when given a low-dimensional random vector of features: $G: z \rightarrow g$. The latter learns to distinguish between the generated images' distribution $p_g=G(p_z)$ and the one from the original examples $p_{data}$. Their objectives can be summarized in the form of a minimax game:
\begin{multline}
    \min_{G}\max_{D}\mathcal{L}_\text{Adv}(G,D)=
    \\
    \mathbb{E}_{x\sim p_{data}}\left[\log D(x)\right] + \mathbb{E}_{z\sim p_z}\left[\log D(G(z))\right].
    \label{eq:adv_loss}
\end{multline}
By providing conditions $c$ (\eg in the form of labels) to both generator and discriminator, the former can learn to synthesize images from a subspace of $p_g$: $G(p_z,c)=p_g(c)\subset p_g$.

Any GAN generator is composed of a series of convolutional modules we label $l_i$. The output of each module, namely $l_k\circ l_{k-1}\circ\dots\circ l_1(z,c)$ constitutes a candidate for an early exit, but it is not a rendered image. For this reason, we need to process it by a series of additional convolutions, before we can retrieve an image from it. These new convolutional $\tilde{l}_i$ modules constitute what we call a \textit{branch}. As portrayed in \cref{fig:pipeline}, we append branches to the backbone architecture after each of its modules. Their depth, \ie the number of modules they are made of, varies in accordance with their attach point. For a backbone built out of $N$ modules, after module $k$, we append a branch of length $N-k$. The branches' modules are less complex, than the backbones', their width, \ie number of channels, is decreased. In this way, at the output of each branch $\tilde{l}_{N}\circ\dots\circ\tilde{l}_{k+1}\circ l_k\circ l_{k-1}\circ\dots\circ l_1(z,c)$, we retrieve an image rendered with a lesser number of computations than at the backbone's output. Each branch is trained by adversarial loss with copies of the backbone original discriminator.

\begin{figure*}
    \centering
    \includegraphics[width=0.99\textwidth]{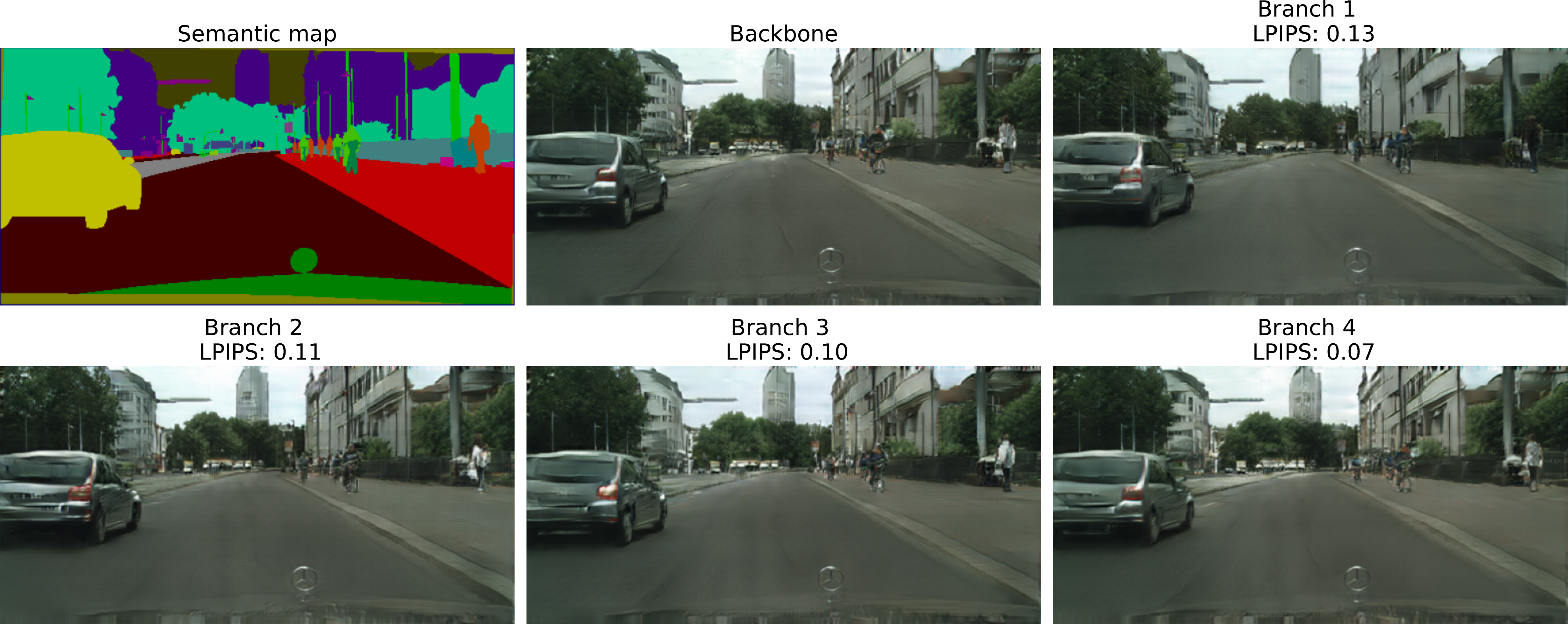}
    \caption{Examples of branches' outputs for the OASIS pipeline. The input consists of a semantic map and a 3D noise.}
    \label{fig:oasis_results}
\end{figure*}

\subsection{Exit predictor}\label{subsec:pred}
During the inference phase, having a set of trained branches, each image can be synthesised through a different exit. Given a quality threshold, we want to be able to select only the branch that will achieve it performing the least possible calculations. To do this, we employ a neural network we call \textit{predictor} $P$, constituted by convolutional and fully connected layers (see supplementary material for details on its architecture and training). We train our predictor by supervised learning, using the backbone inputs $(z,c)$ as training examples, and LPIPS scores $S$ for images generated by branches as labels.
\begin{equation}
    \mathcal{L}_{pred}(z,c;S)= \lVert  P(z,c)-S \rVert^2.
\end{equation}
Once trained, by feeding an input to the trained predictor, we can quickly get an estimation of each branch's output quality, and thus use this information to route the computational flow toward the exit which performs the least computations, while upholding the threshold.


\subsection{Database}\label{subsec:db}
To further improve synthesis quality, we shift from a purely parametric method to a semi-parametric, in which the generating process is guided by patches fetched from a relatively small database. This ensures an increase in quality more prominent in earlier exits, which are the fastest, but suffer the most from the quality decrease due to their lower number of parameters. By adding a moderate amount of memory and computations, we achieve better results, harmonizing the output quality of different branches.

In the database, we store a collection of key-value pairs. Keys are given by applying to the images all the trained layers of the backbone prior to the first branch, and cutting the obtained features into non-overlapping patches. Values are obtained by applying the trained layers of the backbone up to its middle, and cutting the resulting features into patches. During inference, we process each input trough the backbone, up to the layer prior to the first branch. We then take the resulting features, cut them into patches, and for each patch we search the database for the closest key. Once we retrieve the values corresponding to all patches, we glue them together and concatenate the obtained features to the input of each branch.

\subsection{Computational saving metric}\label{subsec:metric}
To quantify the success of our method, we introduce a simple measure of the saved computations.
Since we trade quality for computations, we can use the ratio [saved computations] / [quality loss]. As measure units we will use, respectively, GFLOPs\footnote[1]{Floating point operations} and LPIPS~\cite{Zhang2018lpips}. For instance, in the cross-reenactment of face expressions, we achieve a mean quality gain of $1.3\times10^3$ GFLOPs/LPIPS, meaning that lowering the quality threshold by $+0.01$ LPIPS will yield a decrease of 13 GFLOPs.

\begin{figure*}
  \centering
    \includegraphics[width=0.8\textwidth]{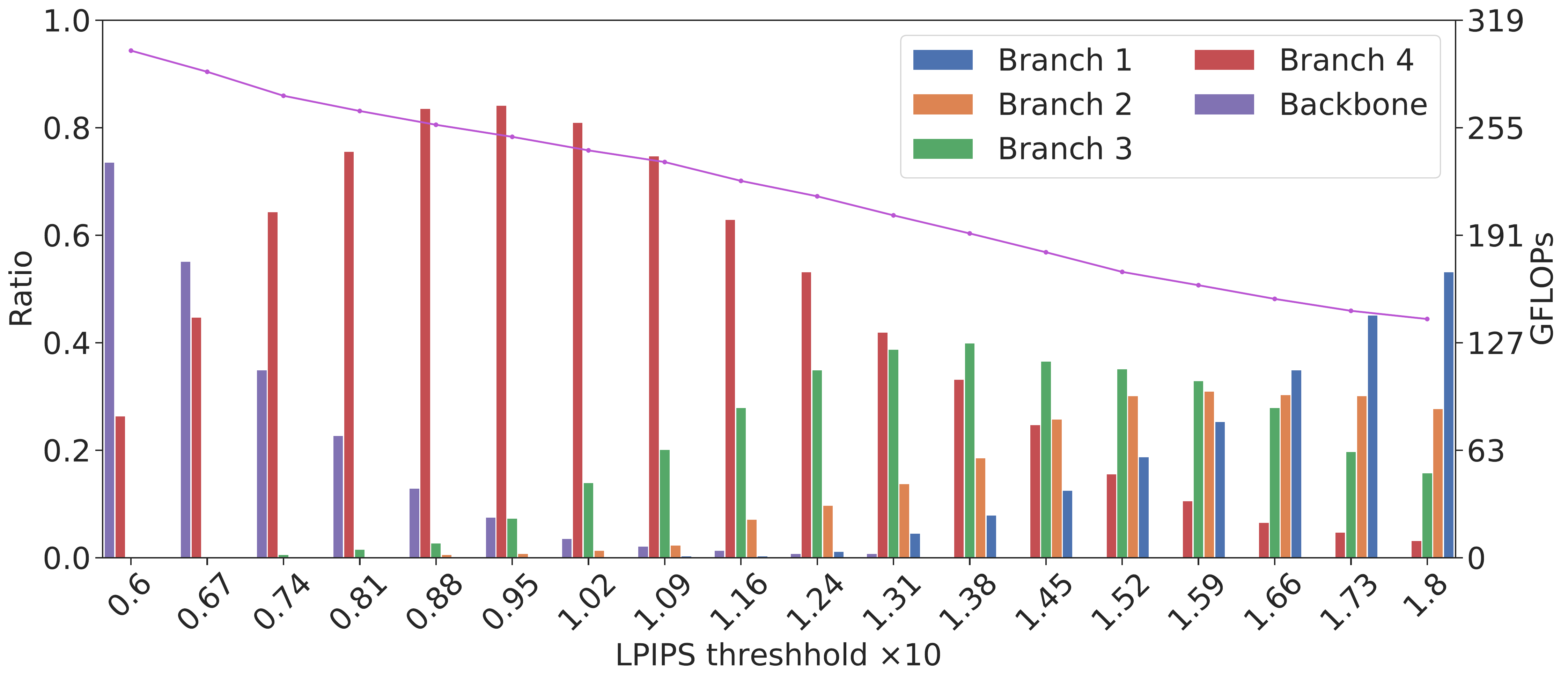}
  \caption{Distribution of computations among branches of the OASIS backbone for a range of imposed LPIPS thresholds. For each threshold, the predictor routes the computation towards one of five possible exits based on the input's complexity it learned.  As quality requirements decrease, the use of the first branches becomes more prominent. All distributions were obtained by sampling the same 500 test images and using SF=$1/4$. Overall GFLOPs for each distribution are shown by the solid line, while their absolute values are shown on the right.}
  \label{fig:oasis_distr}
\end{figure*}

\section{Implementations}
\label{sec:experiments}

\begin{figure} 
    \vskip -0.11in
    \centering
    \includegraphics[width=\columnwidth]{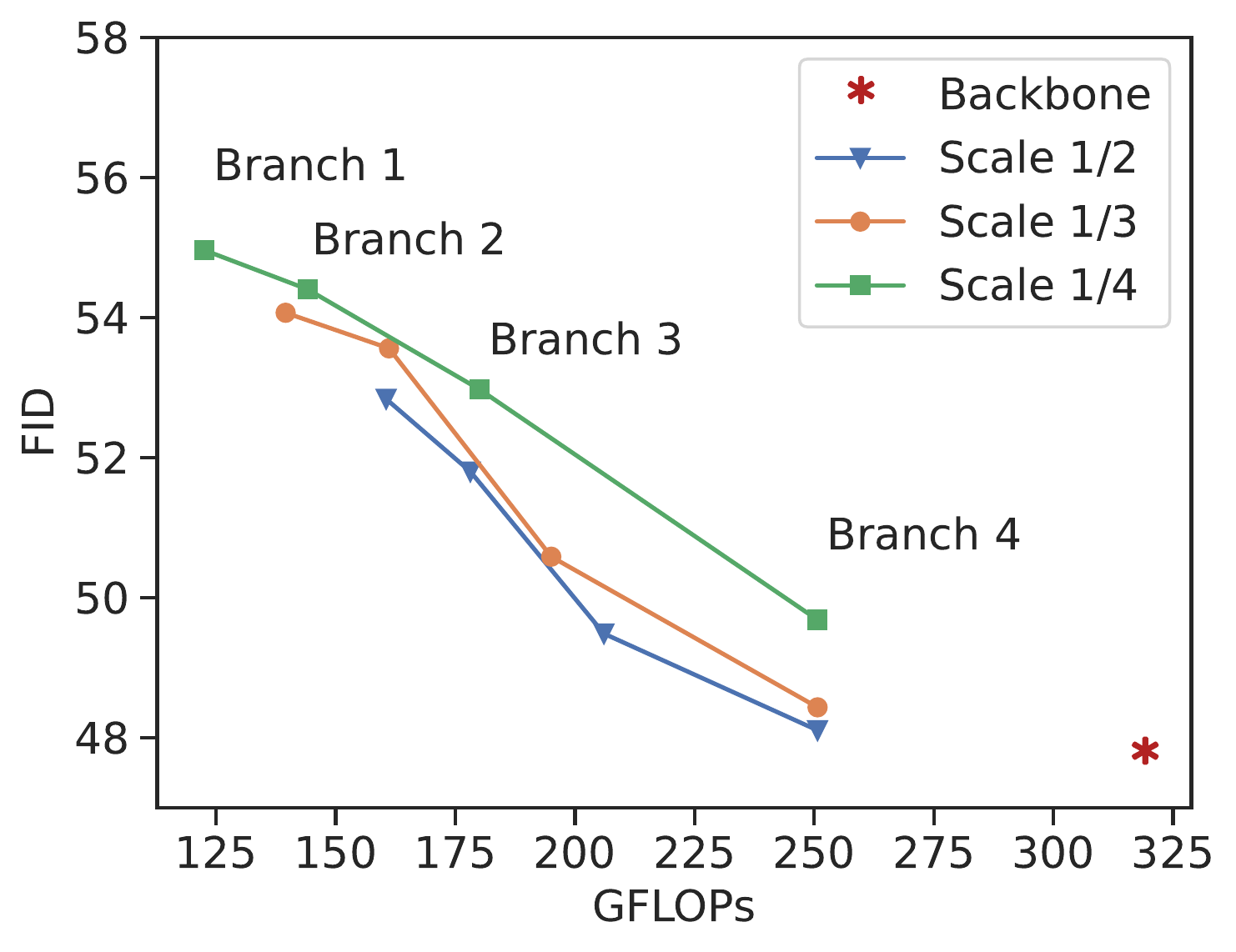}
    \caption{Relation between quality (expressed in FID units) and computations for all branches at different scale factors of the OASIS implementation, with the use of the guiding database.}
    \label{fig:oasis_fvf}
    \vskip -0.15in
\end{figure}

Our method can be applied to a multitude of DNNs for different synthesis tasks. To showcase its generality, we apply it to two distinct image synthesis tasks: (1) Outdoors photographs synthesis starting from a semantic label map, using the Cityscapes dataset~\cite{Cordts2016cityscape}, and taking as backbone the OASIS architecture~\cite{schonfeld2021oasis}; (2) Neural head avatars synthesis, starting from a picture that acts as the avatar's target expression and position, and using as backbone the MegaPortraits architecture~\cite{Drobyshev22megaportraits}.

\subsection{Landscapes from semantic map}

For the implementation of synthesis by semantic map, we used outdoor images with semantic maps from the Cityscapes dataset~\cite{Cordts2016cityscape}. We implemented our pipeline taking as backbone the OASIS model~\cite{schonfeld2021oasis}, which takes as input a semantic map in conjunction with a $3$D noise tensor for diversifying outputs. The OASIS generator consists of $6$ SPADE ResNet modules~\cite{park2019spade}, which in our definitions constitute the backbone modules $l_i, \; i\in[\![1,6]\!]$. 
%
%
We appended $4$ branches, one after each backbone module $l_1$ to $l_4$. The branches' modules $\tilde{l}_i$ were SPADE ResNet modules as well, and their length varied in order to preserve $k + \text{len} = 6 \; \forall k \in [\![1,4]\!]$, as discussed in \cref{subsec:depth_varying_exits}. They constituted a lightweight variant of the backbone modules since we reduced their width, \ie number of channels, by imposing a \textit{scale factor} (SF) $s=1/2,\;1/3,\;1/4$ in order to reduce computations. A detailed explanation of how we scale down channel numbers is given in the supplementary material. We thus created a total of 5 computational routes for each scale factor, their GFLOPs are listed in \Cref{tab:flops_oasis}.
\begin{table}
  \centering
  \begin{tabular}{c c c c c | c}
    \toprule
    SF & 1 & 2 & 3 & 4 & BB\\
    \midrule
    1/2 & 157 & 171 & 193 & 227 & \\
    1/3 & 137 & 154 & 182 & 227 & \textbf{319}\\
    1/4 & 120 & 138 & 168 & 227 & \\
    \bottomrule
  \end{tabular}
  \caption{Comparison between GFLOPs of all 5 computational routes through branches and the OASIS backbone (BB, rightmost column). Different rows correspond to different scale factors (SF). The scale factor does not equally affect all modules, since we imposed a minimum number of channels equal to $64$, after which no further scaling is imposed.}
  \label{tab:flops_oasis}
  \vskip -0.2in
\end{table}

We trained each branch by imposing adversarial losses, as in \cref{eq:adv_loss}, generated by competing against copies of the OASIS discriminator. Alongside, we also imposed VGG~\cite{johnson2016vgg} and LPIPS~\cite{Zhang2018lpips} losses using as ground truth the image synthesized by the backbone.
\begin{equation}
    \mathcal{L}_\text{Branch} = \mathcal{L}_\text{OASIS} + \alpha\mathcal{L}_\text{VGG} + \beta\mathcal{L}_\text{LPIPS},
    \label{eq:oasis_branch_loss}
\end{equation}
where $\alpha$ and $\beta$ are hyperparameters we chose in order to equalize the losses' contribution. A thorough list of all hyperparameters and training details is given in the supplementary material.

\begin{figure*}
    \centering
    \includegraphics[width=0.99\textwidth]{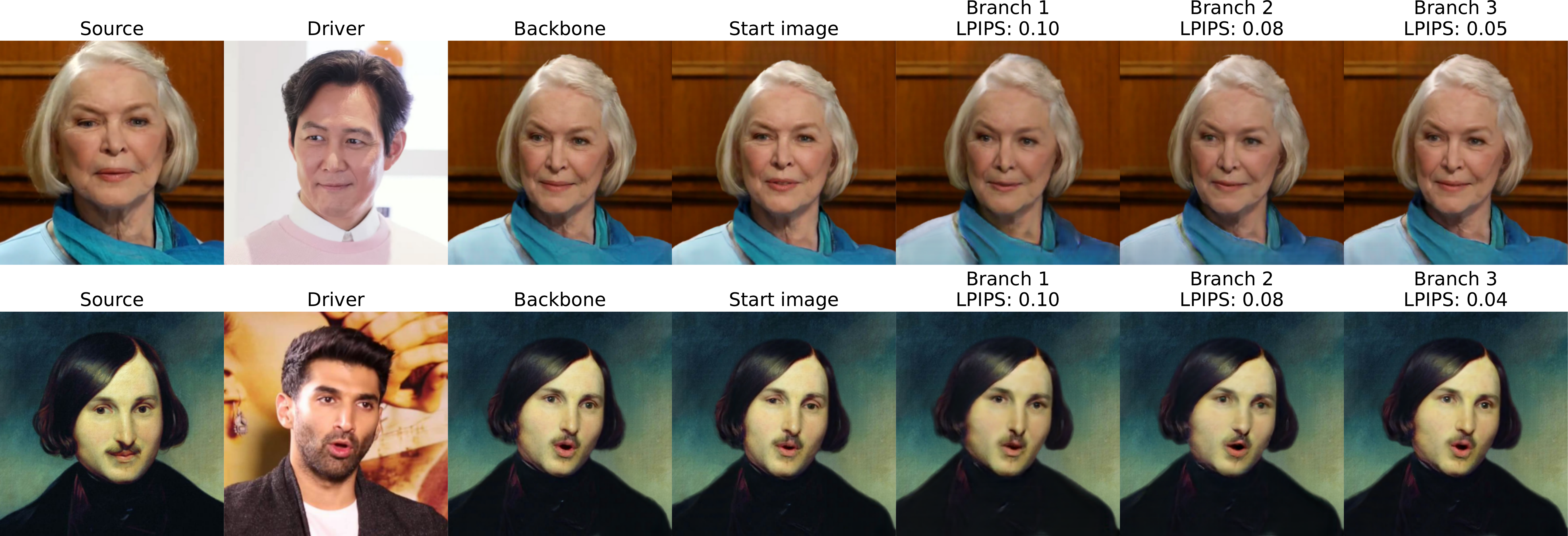}
    \caption{Examples of branches' outputs for the MegaPortraits pipeline. The model has a source face whose traits must be transferred to the driver's face. The synthesis is guided by the start image, which was selected from a database containing 960 frames with the source appearance, because of its similarity to the driver's pose and expression. LPIPS is measured cutting out the background.}
    \label{fig:avatars_results}
\end{figure*}

In order to implement the key-value database for guiding image generation~(as discussed in \cref{subsec:db}), we randomly selected $500$ semantic maps from the train dataset. For each one of them, we created $100$ different inputs using a fixed set of $3$D noises. We fed the inputs into the first $2$D convolutional layer and the subsequent ResNet module of the backbone. The obtained features were then divided into $8\times 16=128$ non-overlapping patches, in accordance with their resolution, which gave us the keys. The values were extracted by processing the inputs up to the third ResNet module of the backbone and cutting the obtained features into the same patches. The database is populated once at the beginning of the training phase. To decrease the redundancy in keys, we applied FPS sampling~\cite{eldar1997fps} to them.
During the forward phase, after an input was processed through the first $2$D convolutional layer and the subsequent ResNet layer, it was divided into $128$ patches. Subsequently, the database was searched for the key most similar to each patch with the aid of the FAISS library~\cite{johnson2019faiss}. All $128$ retrieved values were then glued accordingly. We used this composed feature to guide the synthesis process by concatenating it to each branch's input after due resizing performed by a convolution. The resulting distribution of quality among all branches, evaluated by the Fréchet inception distance (FID)~\cite{heusel2017fid}, is shown in \cref{fig:oasis_fvf}.
%


Finally, the pipeline comprehending all generating branches and the backbone, together with the database guidance, was used to produce the dataset for training the predictor (as discussed in \cref{subsec:pred}). Since the OASIS input consists of a semantic map and a high-dimensional random noise space, we restricted the training to $100$ fixed noise vectors in combination with the Cityscapes train set. In this way we achieved a mean error of 5\% on the validation set.

The overall result for the whole pipeline at SF$=1/4$ is summarized by \cref{fig:oasis_distr}. The latter shows the distribution of branches chosen by the predictor at various quality thresholds. One can see how different thresholds affect the exit's choice: while imposing very high quality narrows the spectrum of possible exits, at lower (but nonetheless high) requirements, all additional branches are utilized. Most importantly, the GFLOPs count shows a dramatic decrease of computations when earlier branches are used. By approximating the GFLOPs curve to a constant slope, we can estimate a mean gain factor of $1.2\times 10^3$ GFLOPs/LPIPS.

\subsection{Neural head avatars}

For the neural head avatar implementation, we exploited the VoxCeleb2 dataset \cite{chung2018voxceleb2}. 
\begin{table}
  \centering
  \begin{tabular}{c c c c | c}
    \toprule
    SF & 1 & 2 & 3 &  BB\\
    \midrule
    1/3   & 65  & 100 & 136  &\\
    1/6   & 51  & 89 & 135  &\textbf{213}\\
    1/15   & 47  & 85 & 127  & \\
    \bottomrule
  \end{tabular}
  \caption{Comparison between GFLOPs of all 4 computational routes through branches and the MegaPortraits backbone (BB, rightmost column). Different rows correspond to different scale factors (SF).
  }
  \label{tab:head_flops}
  \vskip -0.15in
\end{table}
We based ourselves on the MegaPortraits generating method~\cite{Drobyshev22megaportraits} for $512\times512$~pixels images. This pipeline consists of multiple steps ensuring the transfer of traits from a source face to a driver face. We took as backbone modules $l_i, \; i\in[\![1,9]\!]$ its final set of modules comprehending $9$ residual blocks, which amount to a total of $213$ GFLOPs. We attached $3$ branches, one after backbone's block number $2,\; 4,$ and $6$. Their modules $\tilde{l}_i$ were the same residual blocks, and their respective depth, \ie number of modules, mirrored that of the remaining path: $8,\; 6$ and $4$, thus maintaining $k + \text{len} = 9 \; \forall k \in \{2,4,6\}$.
To lighten the branches, we imposed three different scale factors to the modules' width, \ie number of channels. Their overall GFLOPs are listed in \cref{tab:head_flops}.

We trained our branches by imposing adversarial losses, as in \cref{eq:adv_loss}, obtained competing with copies of the MegaPortraits discriminator. Alongside, we imposed VGG~\cite{johnson2016vgg}, MS-SSIM~\cite{wang2003mssim} and $\mathcal{L}_1$ losses between the branches' and the backbones' synthetic images. Additionally, we used the backbone's intermediate features to impose a feature-matching loss~(FM)~\cite{wang2018feature_matching} and retained the original gaze loss~(GL)~\cite{Drobyshev22megaportraits}.
\begin{multline}
    \mathcal{L}_\text{Branch} = 
    \\
    \mathcal{L}_\text{Adv} + c_1\mathcal{L}_\text{VGG} + c_2\mathcal{L}_\text{MS-SSIM} + c_3\mathcal{L}_\text{1} + c_4\mathcal{L}_\text{FM} + c_5\mathcal{L}_\text{GL},
    \label{eq:head_branch_loss}
\end{multline}
where coefficients $c_i$ were chosen to harmonize the losses' effects. A list of all hyperparameters and training details is given in the supplementary material. 

\begin{figure}
    \centering
    \includegraphics[width=\columnwidth]{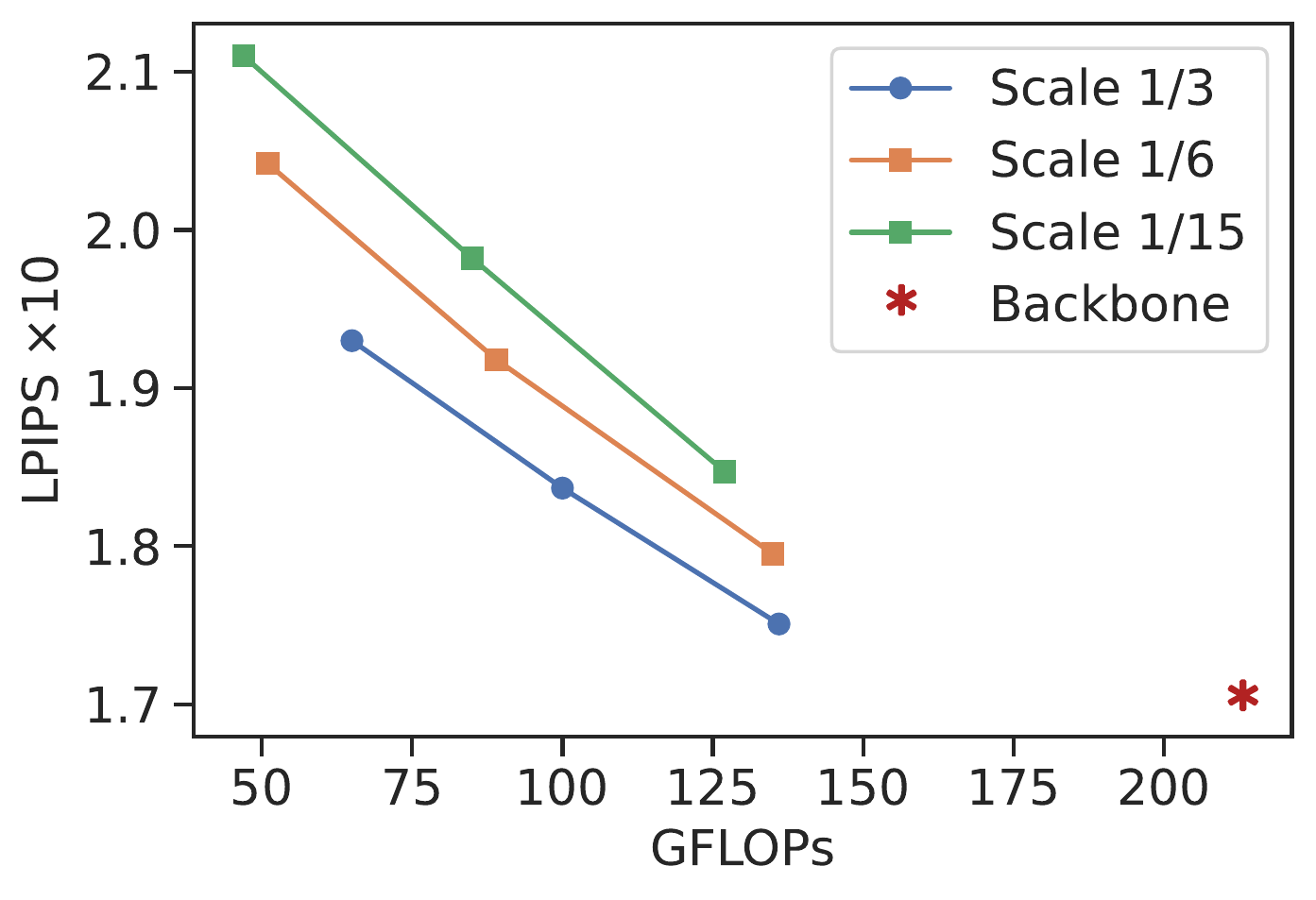}
    \caption{Relation between quality (expressed in LPIPS units) and computations for all branches at different scale factors of the MegaPortraits implementation, with the use of the guiding database.}
    \label{fig:head_fvf}
    \vskip -0.15in
\end{figure}

We populated our database by pictures of the source face with a plethora of different orientations and expressions. At each iteration, we searched the database for the face most similar to the driver's, \ie the one which orientation and expression we want to obtain. To perform this search, we fed the driver to the first module of MegaPortraits, which extrapolates the angles describing face direction, and a multi-dimensional vector which encodes face expression. We exploited this feature and designed a different key-value search. We employed 3 angles for the encoding of face directions, while the expression space is $512$-dimensional.
Once we obtained a key characterizing the driver, we looked for the closest one from the images in the database. The retrieved value was then concatenated to the input of each branch module $\tilde{l}_i$ after due resizing.
The resulting distribution of quality among branches is shown in \cref{fig:head_fvf}.
%

\begin{figure}
    \centering
    \includegraphics[width=\columnwidth]{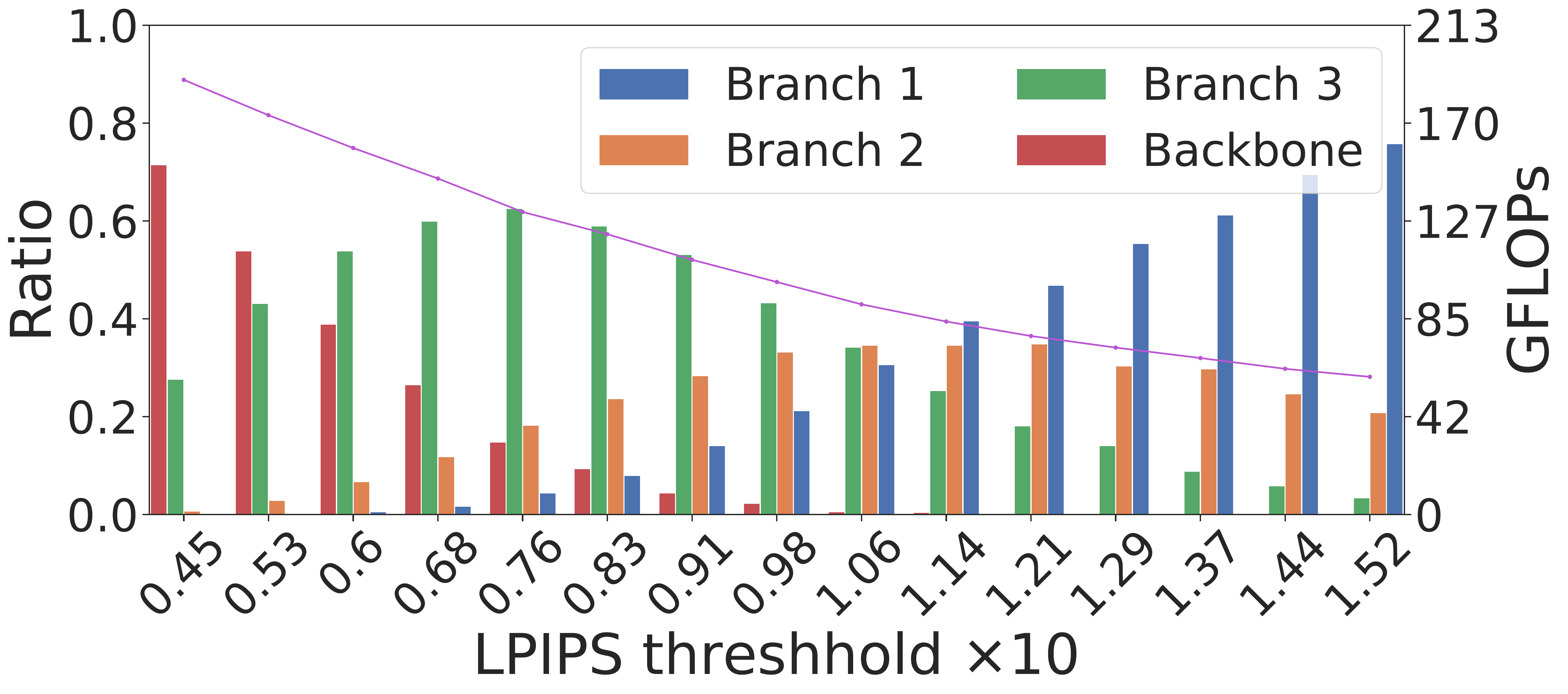}
    \caption{Distribution of computations among branches of the MegaPortraits backbone for a range of imposed LPIPS thresholds. For each threshold, the predictor routes the computation towards one of four possible exits based on the input's complexity it learned. All distributions were obtained by sampling the same 702 test images and using SF=$1/15$. Overall GFLOPs for each distribution are shown by the solid line, while their absolute values are shown on the right.}
    \label{fig:head_distr}
    \vskip-0.2in
\end{figure}

Finally, we trained the predictor as discussed in \cref{subsec:pred}, on LPIPS scores obtained comparing the branches' with the backbone's output. Afterwards, we were able to impose any quality threshold and the predictor was able to choose the path that satisfied it with the least computation. The overall results for the whole pipeline are summarized by \cref{fig:head_distr}. One can see how lower-quality thresholds can be maintained with a great decrease in GFLOPs due to the use of lighter branches. By approximating the GFLOPs curve to a constant slope, we can estimate a mean gain factor of $1.3\times 10^3$~GFLOPs/LPIPS.

\begin{figure}
    \centering
    \includegraphics[width=0.9\columnwidth]{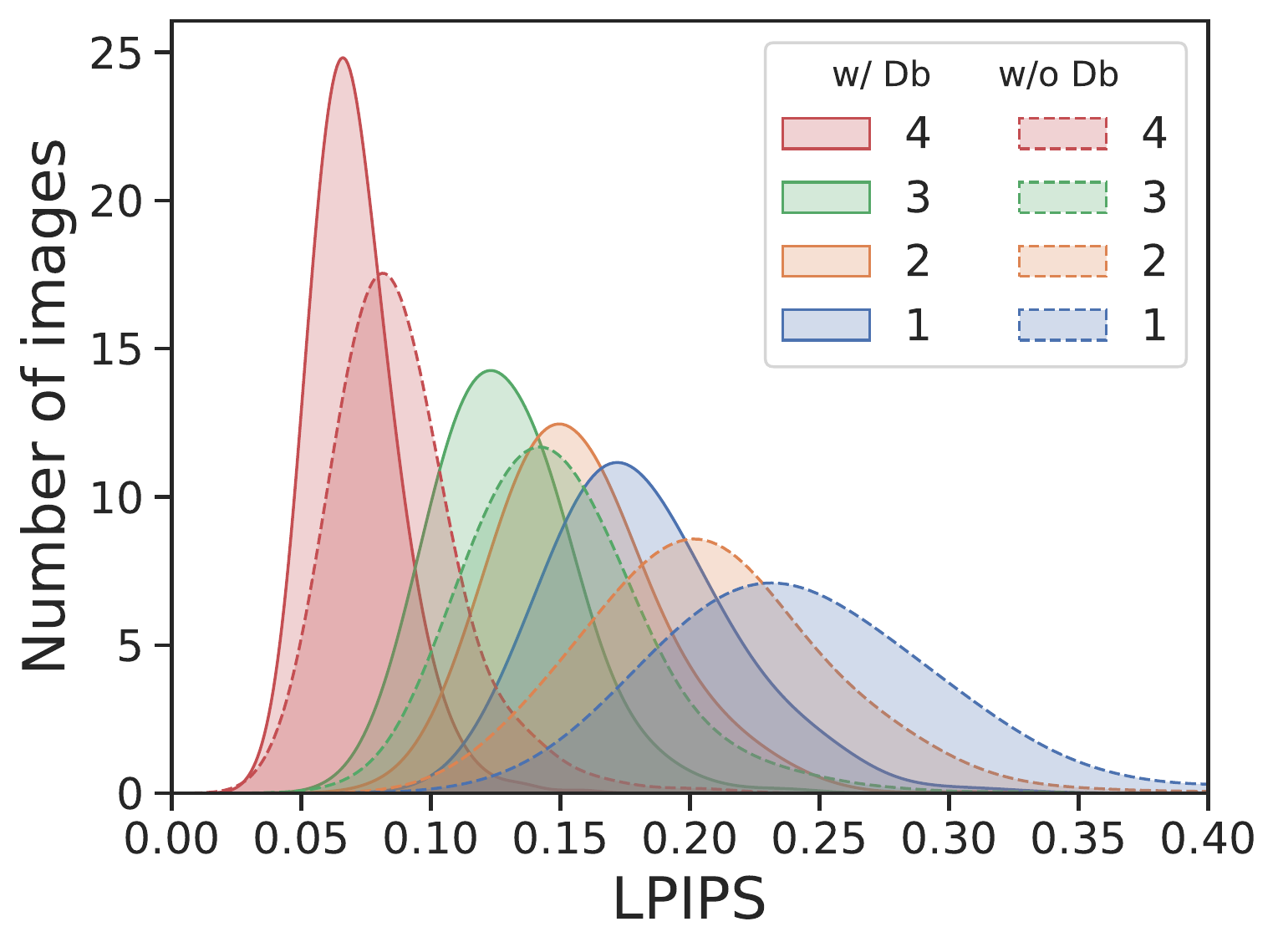}
    \caption{Comparison between quality distributions for the OASIS pipeline with SF=$1/4$, with the use of the guiding database and without it. LPIPS were obtained by comparison with backbones' images. Curves were drawn sampling 500 images and applying kernel density estimation with bandwidth $0.3$.}
    \label{fig:db_distr}
\end{figure}

\begin{figure}
    \centering
    \includegraphics[width=\columnwidth]{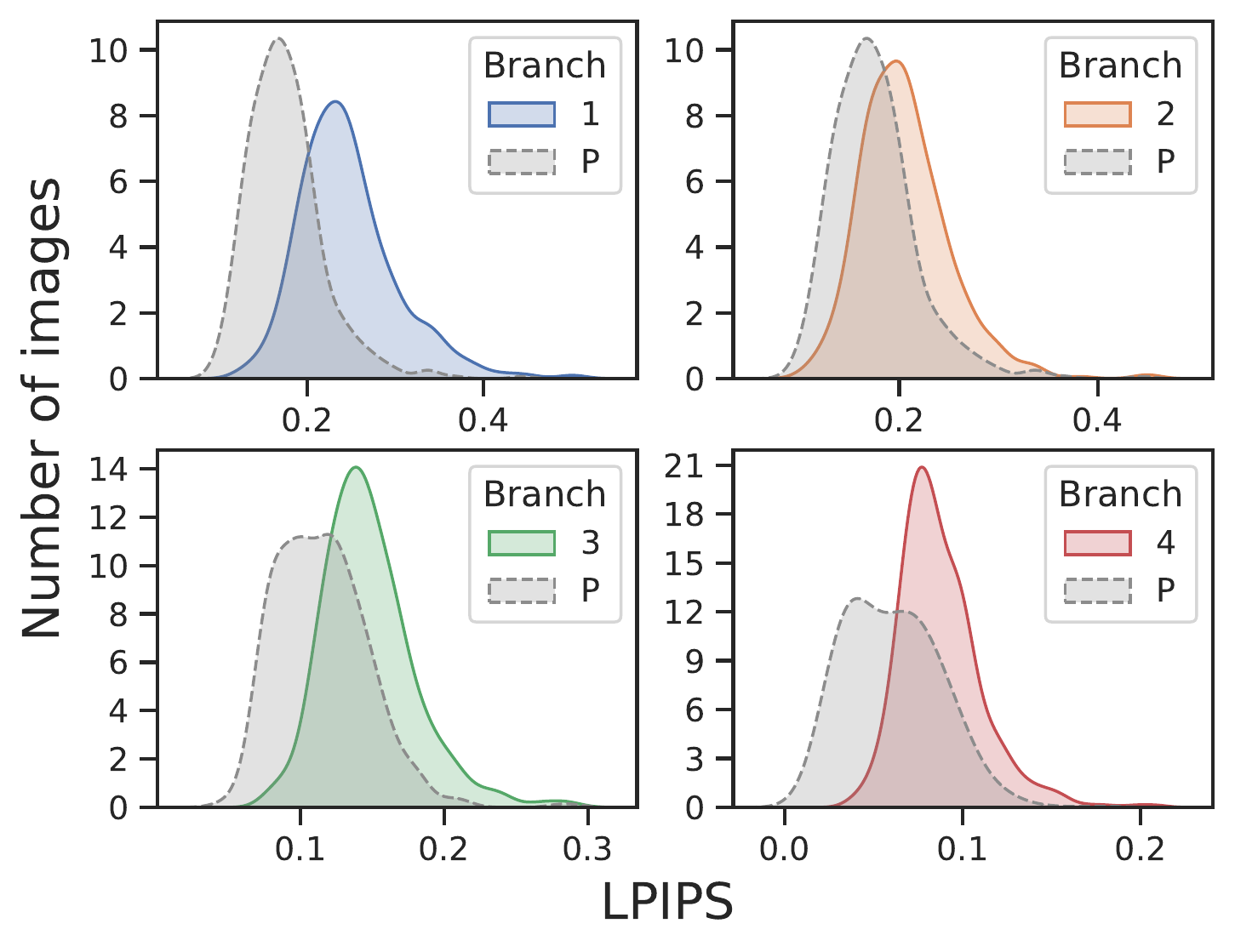}
    \caption{Comparison between quality distributions of single OASIS branches, and quality distributions obtained by use of the predictor (P). The predictor was set to enforce thresholds equal to the branches' mean quality. LPIPS were obtained by comparing images of branches for SF=$1/4$ with backbones' images. Each distribution is sampled by inputting 500 semantic maps with random noises. The curves are the result of kernel density estimation with bandwidth $0.3$.}
    \label{fig:oasis_densities}
    \vskip -0.2in
\end{figure}

\section{Ablation study}

Although the image generation is possible without the database of guiding images, we find it essential for ensuring the quality of earlier branches. It can be in fact argued that its implementation harmonizes exits' output quality, by affecting the most the earliest branches, as testified by \cref{fig:db_distr}.

Additionally, the database can be used to amend for the deficiency of the training set. As we will see in the next section, part of the difficulty in rendering is due to a lack of the DNN training, which may very well be inherent to the specific task, as for neural head avatars generation. By providing guiding examples, we somewhat ``patch" the holes in the training.

As discussed, our implementation of the dynamical routing relies on the creation of suitable early exit, as well as the use of a predictor. The latter is essential to enforce custom quality thresholds, since the use of single exits is will produce only images with fixed quality distributions.

Furthermore, although all branches have a certain mean quality, captured by their FIDs (see \cref{fig:oasis_fvf} as an example), we can't rely on just a single branch to produce images with consistent quality. The variation in quality of each exit is quite wide and it gets wider in the earliest ones, as portrayed in \cref{fig:db_distr}. The predictor prevents this by choosing a heavier branch when quality can't be provided by a lighter one. The comparison between quality distributions of images obtained from single branches and those obtained by the use of the predictor, set to output a threshold equal to the branches' mean quality, is shown in \cref{fig:oasis_densities}. We can clearly see how the predictor enforces the threshold by routing difficult images towards the next branches, thus shifting the distribution.


\section{Discussion}

Our method is widely applicable, since it can be applied to all models that employ a multi-layer decoder, as illustrated in \cref{fig:pipeline}. The presence of multiple layers is our only requirement, since branches take as input the output of these layers. This
includes models that take random noise as input, such as StyleGAN. Such implementation is almost identical to the one for the OASIS model, only without the concatenation of a semantic map to the noises used in the database and for the predictor training.

\begin{figure}
    \centering
    \includegraphics[width=0.8\columnwidth]{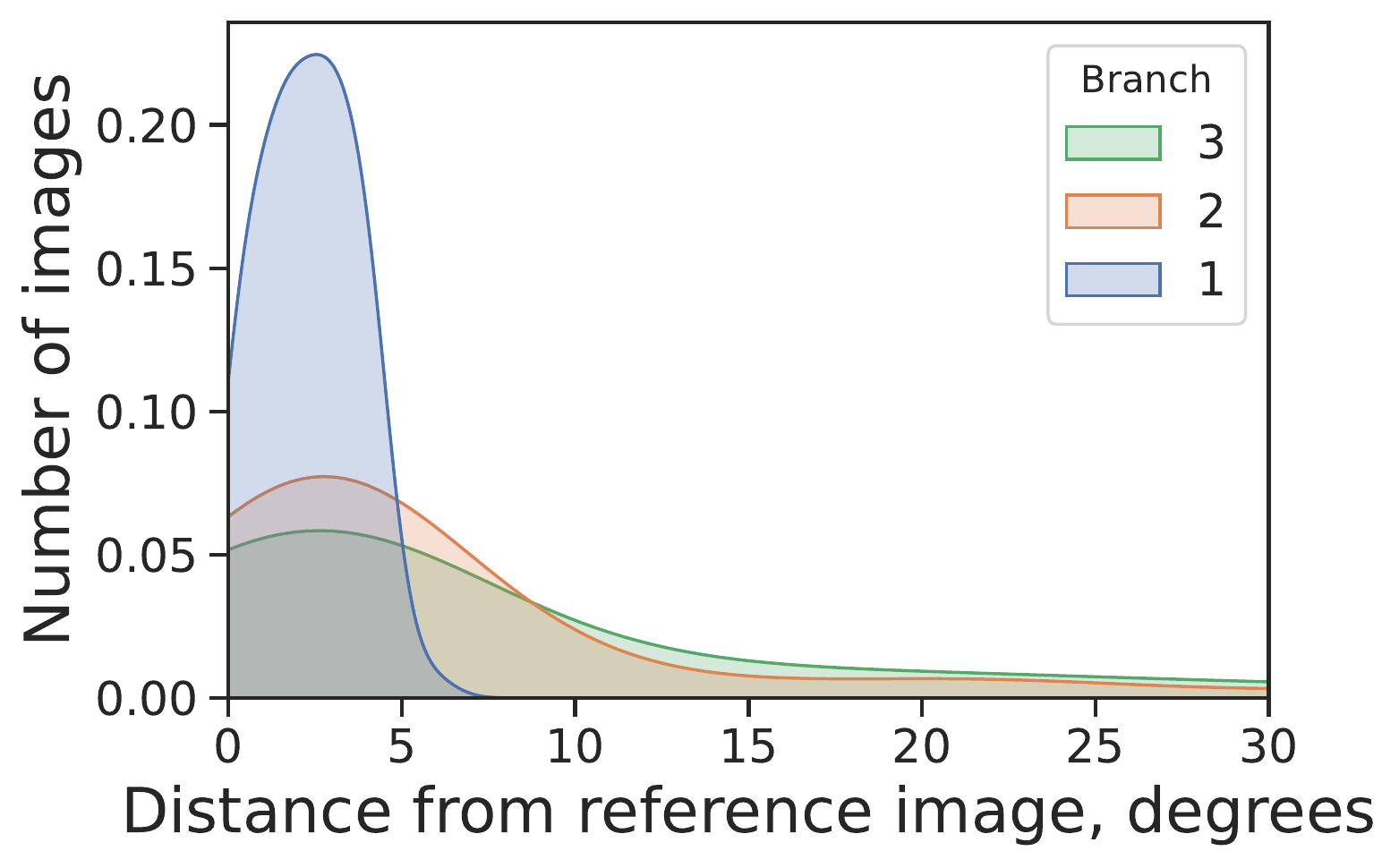}
    \caption{Comparison between number of images routed to different branches in relation to their head rotation. We used the branches for SF=$1/15$. Distributions were obtained by sampling $702$ images in total. Curves are the result of kernel density estimation with bandwidth $0.5$. The quality threshold was set at $0.09$ LPIPS.}
    \label{fig:distance}
    \vskip -0.2in
\end{figure}

As we stated, not all images are equally difficult to generate. This irregularity lays at the core of our method. A multitude of reasons is responsible for such uneven difficulty distribution. For instance, if we consider the neural head avatar generation problem, one may argue that the DNN is not ideally trained. Some head rotations or expressions may be less present during the training phase, and thus require a heavier model to output images with high quality. We analyzed this problem by comparing images with different head rotations and expressions, and their quality. Specifically, by using our pipeline, we generated $702$ head avatars and looked at which branch they were routed by the predictor. By plotting the faces distribution in relation to the angle between them and database images used for guidance, we could clearly see how the rendering difficulty is correlated with this distance. The greater the angle between the two images, the higher the difficulty gets, as reported in \cref{fig:distance}.

\section{Limitations}

Although our method can save a great amount of computations, it has some limitations. One can not apply our pipeline as it is to transformers and other synthesis algorithms that don't comprehend a decoder. There is no single recipe for populating the database. We chose to populate it randomly, but this may actually not be the best choice.
Since we need to generate a training dataset for the predictor, we need additional training inputs, thus the size of viable databases is increased.
All the branches need additional training, and the memory used for storing the whole pipeline is higher that the one used for the original DNN.
We assume the batch size is equal to 1 due to real-time application scenario. In case of batched execution, it will require
some orchestration and batch accumulation algorithm in order to infer efficiently, since different images in batch may
undergo different computational routes.




{\small
\bibliographystyle{my_ieee.bst}
\bibliography{biblio.bib}
}

\pagebreak
\begin{center}
\textbf{\large Supplementary material for FIANCEE: Faster Inference of Adversarial Networks via Conditional Early Exits}
\end{center}

\setcounter{equation}{0}
\setcounter{figure}{0}
\setcounter{table}{0}
\setcounter{page}{1}
\makeatletter
\renewcommand{\theequation}{S\arabic{equation}}
\renewcommand{\thefigure}{S\arabic{figure}}
\renewcommand{\thetable}{S\arabic{table}}



\newcommand{\Var}{\mathrm{Var}}
\renewcommand{\thesection}{S\arabic{section}}
\setcounter{section}{0}
\section{Qualitative results}
\subsection{The OASIS pipeline}
\subsubsection{Architectures and dimensions}

\begin{table*}
  \centering
  \begin{tabular}{c c |c c| c c| c c | c c}
    \toprule
    \textbf{SF} & \textbf{Bank}  & \multicolumn{2}{c|}{\textbf{Branch 1}} & \multicolumn{2}{c|}{\textbf{Branch 2}} & \multicolumn{2}{c|}{\textbf{Branch 3}} & \multicolumn{2}{c}{\textbf{Branch 4}}  \\
     & & FID$\downarrow$ & mIOU$\uparrow$ & FID$\downarrow$ & mIOU$\uparrow$ & FID$\downarrow$ & mIOU$\uparrow$ & FID$\downarrow$ & mIOU$\uparrow$  \\
     
    \midrule
    \multirow{2}{*}{1/2} & \xmark & 64.2 & 59.8 & 59.3 & 62.6& 55.9 & 62.2& 50.1 & 64.2 \\
     & \cmark & 52.8 & 67.5& 51.8 &  68.7& 49.5 & 68.5& 48.1 &69.3 \\ [0.1cm]
    \multirow{2}{*}{1/3} & \xmark & 65.9 & 61.4 & 59.5 & 61.6 & 57.2 & 65.2 & 53.1 & 69.4   \\
     & \cmark & 54.1 & 65.5 & 53.6 & 69.6 & 50.6 & 68.8 & 48.4 & 69.6   \\ [0.1cm]
    \multirow{2}{*}{1/4} & \xmark & 69.6 & 57.5 & 62.2 & 61.8 & 56.4 & 65.5 & 53.0 & 68.3  \\
      & \cmark & 54.9 & 65.5 & 54.4 & 67.1 & 53.0 & 66.7 & 49.7 & 69.4  \\
    \midrule
    \multicolumn{2}{c}{Backbone}&
    \multicolumn{2}{c}{} & \multicolumn{2}{c}{} & \multicolumn{2}{c}{} & 47.7 & 69.3  \\
    \bottomrule
  \end{tabular}
  \caption{Quantitative results for the OASIS pipeline. The minimum number of channels is 64.}
  \label{tab:metrics_oasis_fid_miou_min64}
  \vskip -0.05in
\end{table*}

\begin{figure*}
    \centering
    \includegraphics[width=0.33\textwidth]{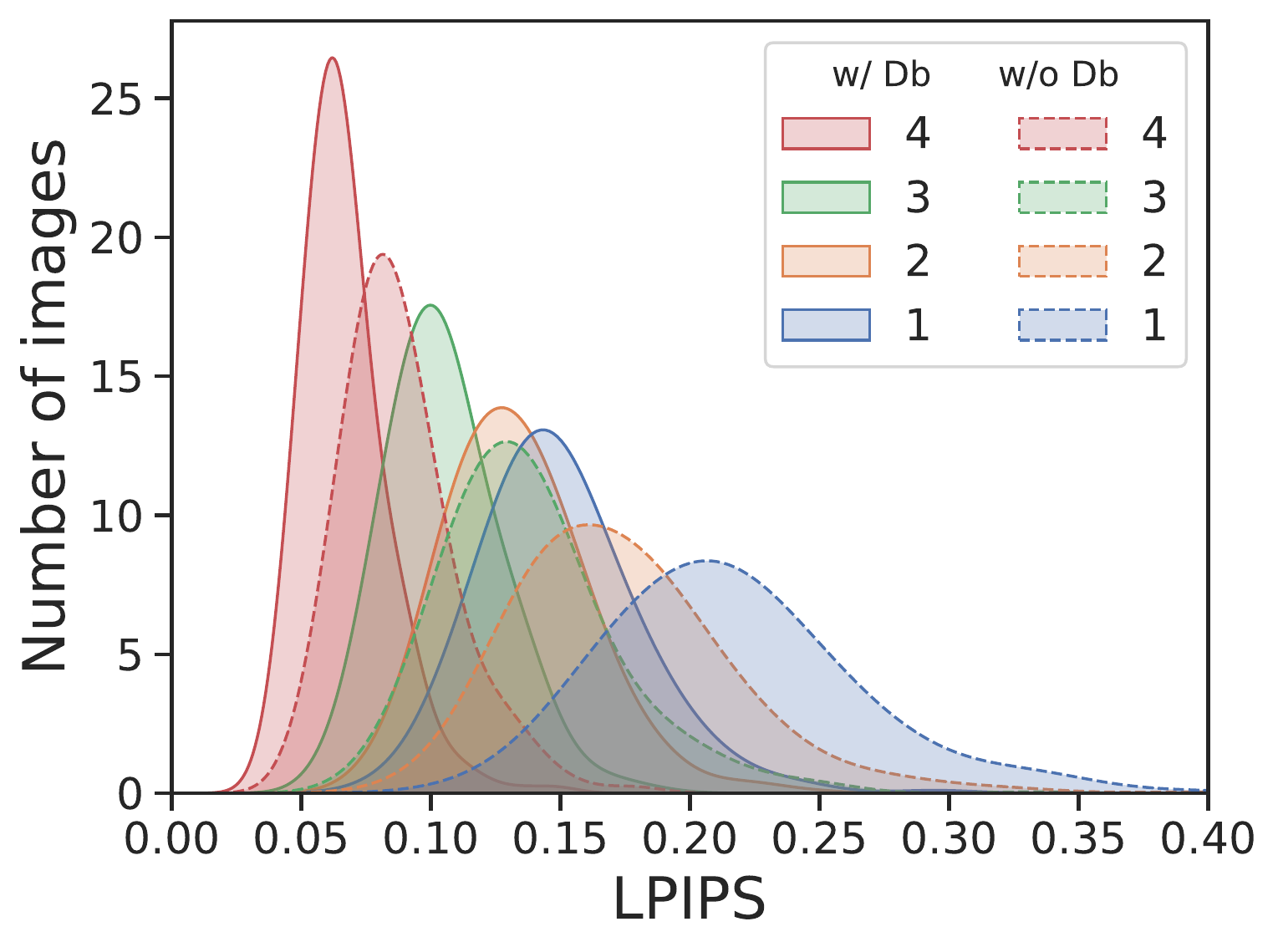}
    \includegraphics[width=0.33\textwidth]{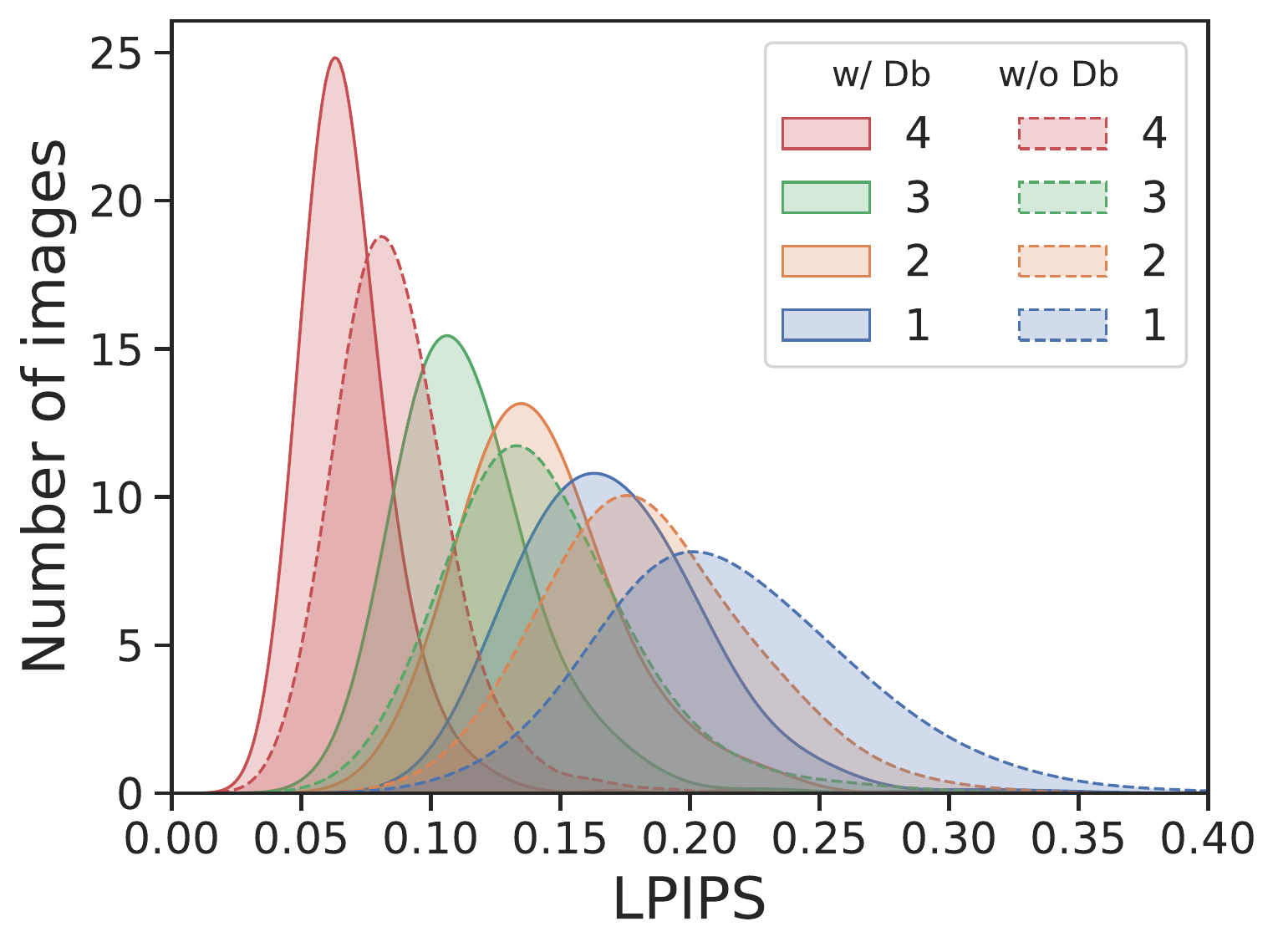}
    \includegraphics[width=0.33\textwidth]{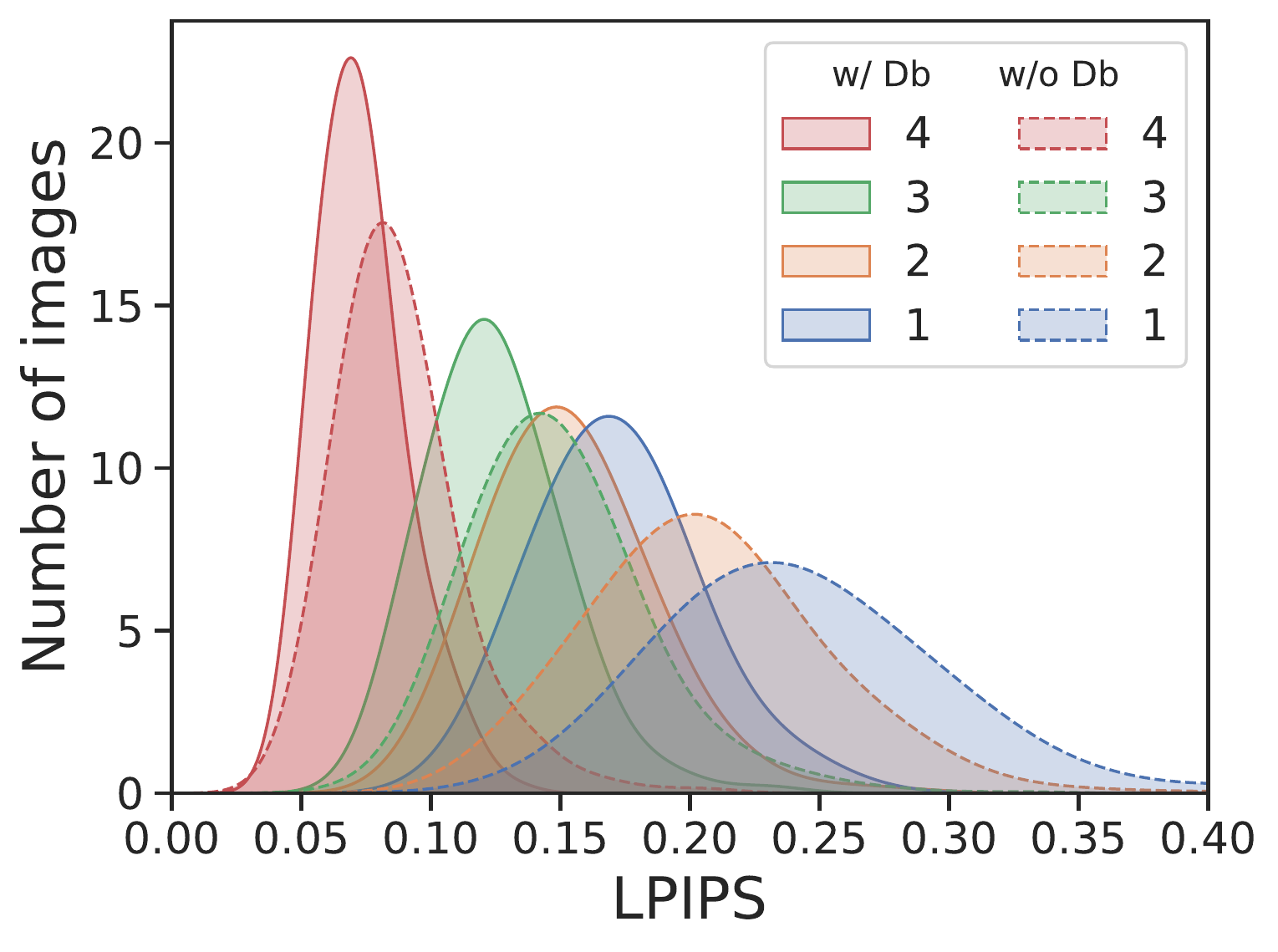}
    \caption{OASIS pipeline, comparison between the database effect to quality distribution for different scale factors. The minimum number of channels is 64. Left to right: SF $=1/2,\;1/3,\;1/4$.}
    \label{fig:OASIS_db}
\end{figure*}

The original OASIS \cite{schonfeld2021oasis} generative DNN consists of an initial $2$D convolutional layer, followed by 6 SPADE-ResBlock modules \cite{park2019spade} and a final Conv2D, LeakyRelu, and TanH. Its total number of parameters is 74M. We appended $4$ branches to it, after ResBlock $1,\;2,\;3$, and $4$ respectively. Each branch consisted of the same number of ResBlock modules as the remaining part of the backbone. In order to create lighter computational paths, we decreased the number of channels of the branches' modules. To do it in a coherent manner, we decided to scale down all channels uniformly by multiplying them by a \textit{scale factor} (SF). Since such scaling with arbitrary coefficients may produce channel numbers too small to be of use, we restrained its effect by imposing a minimum number of channels, under which no scaling was forced. In other words, if the minimum number is 64, and we enforce factor of 1/3 starting from 128 , the new channel number will be 64, instead of 43. We explored a plethora of different scale factors and minimum channels, which we report in \Cref{tab:OASIS_m64}. 
The database we employed was created using $500$ semantic maps randomly chosen from the training dataset, each concatenated with $100$ different $3$D noise tensors to produce a variety of inputs, that were processed and divided into $128$ non-overlapping patches, yielding a total of 
$500\times100\times128=6.4\text{M}$ key-value pairs. Since redundancy in the key space is rather probable, we extracted from this multitude of pairs only up to $5$K for each semantic class using FPS sampling~\cite{eldar1997fps}, for a total of $122\,100$ pairs. Each key is a $1024$-dimensional vector, and each value consists of a float32 tensor of dimensions $(512,4,4)$. The total size of stored parameters is thus $1.1$G.

During the retrieval, the guiding features are taken after the first Conv2D and ResNet blocks of the backbone. Then, for each on the $N\in[1,35]$ semantic classes present in the input, these features are cut into $128$ patches and their $1024$-dimensional space is scanned in order to find the closest key from the database with corresponding semantic class. This search is performed quite rapidly thanks to the FAISS library~\cite{johnson2019faiss}, and thus does not burden computations.

Once retrieved all $128$ patches, a guiding feature is constructed by gluing them together. This feature is concatenated to the input of each branch, and for this reason their number of channels must be increased. When employing the database, the input channels for the first ResBlocks in each branch, reported in \cref{tab:OASIS_m64}, are multiplied by 1.5.
The memory overhead from the MegaPortrait’s database amounts to $21$ Mb. We expect it to be uploaded to the GPU memory, since its size is much lower than that of the network, weighting $131$ Mb. On a desktop GPU (P$40$), the retrieval latency is up to $5$ ms, which constitutes $16\%$ of the smallest branch’s inference time. We assume our method will be used to speed up networks used in real-time applications, which usually run on edge devices, and use a batch of size 1 to minimize latency. 

The last key component of our pipeline is the Predictor. It's architecture is summarized in \Cref{tab:avatar_pred}.
\begin{figure}[H]
    \centering
    \includegraphics[width=0.45\textwidth]{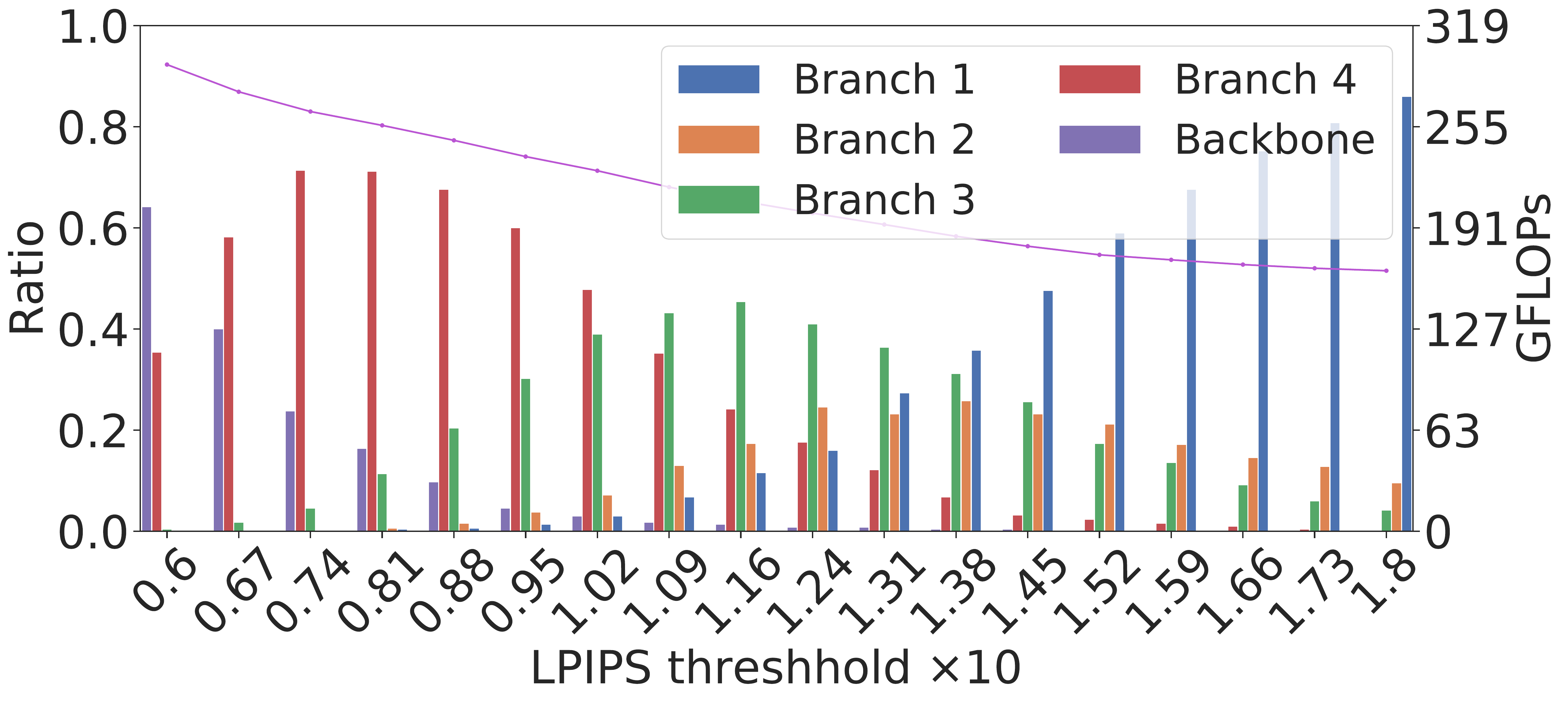}
    \includegraphics[width=0.45\textwidth]{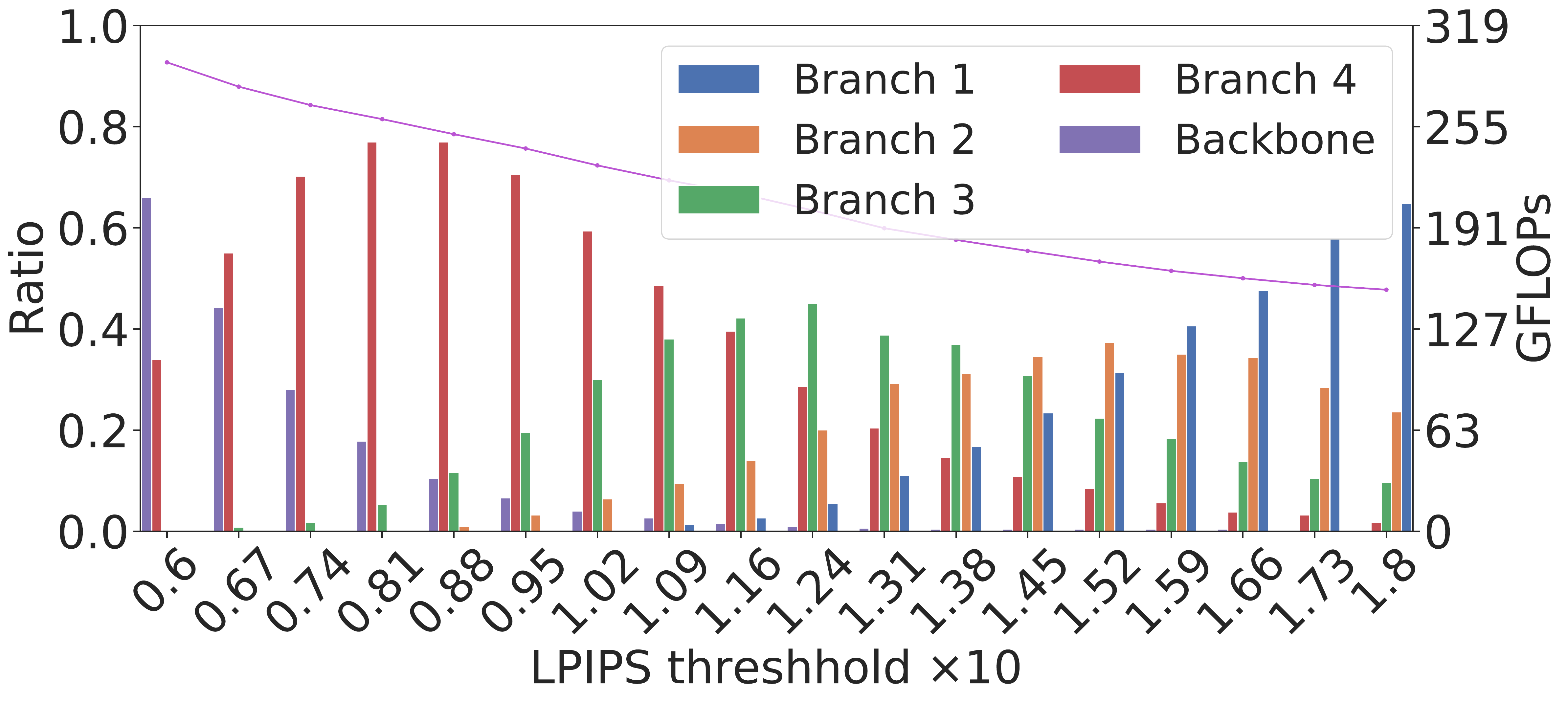}
    \includegraphics[width=0.45\textwidth]{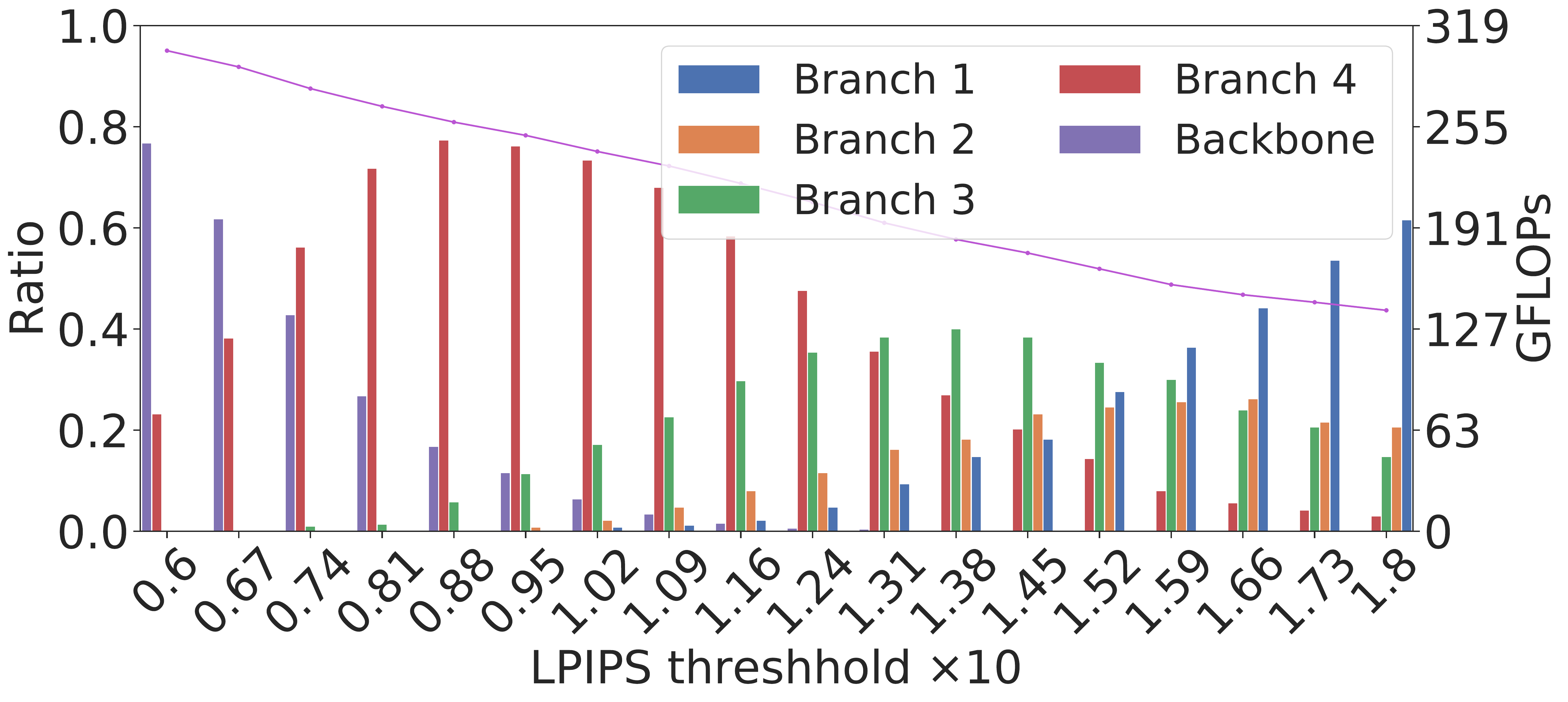}
    \caption{OASIS pipeline, comparison between the efficacy of different scale factors. The minimum number of channels is 64. From top to bottom: SF $=1/2,\;1/3,\;1/4$.}
    \label{fig:OASIS_tzars}
\end{figure}
\subsubsection{Training details}

For the implementation of our method we had to train all branches and the predictor.

Branches for OASIS were trained by competing against copies of
the original OASIS discriminator. Alongside, we also imposed
VGG \cite{johnson2016vgg} and LPIPS \cite{Zhang2018lpips} losses using as ground truth the image synthesized by the backbone,

\begin{equation}
    \mathcal{L}_\text{Branch} = \mathcal{L}_\text{OASIS} + \alpha\mathcal{L}_\text{VGG} + \beta\mathcal{L}_\text{LPIPS},
    \label{eq:oasis_branch_loss}
\end{equation}
where the overall learning rate was set to $4\times10^{-4}$ and the coefficients were set to $\alpha=10$ and $\beta=5$ in order to equalize the losses' contribution. The discriminators retained their original losses. Both the generator and the discriminators were trained via Adam optimization~\cite{kingma2014adam} with $\beta_1 = 0$, $\beta_2=0.999$.
The computations were performed using distributed data parallel from the PyTorch library~\cite{paszke2019pytorch} onto 2 P40 NVIDIA GPUs with batch $=2$ and lasted approximately $6$ days. The resultant qualities can be found in \cref{tab:metrics_oasis_fid_miou_min64} and \cref{tab:metrics_oasis_fid_miou_min32}.

The OASIS predictor was trained to output images' quality for each branch. We did it by imposing minimum squared error loss between its predictions and the actual qualities:
\begin{equation}
    \mathcal{L}_{\text{Pred}}(z,c;S)= \lVert  P(z,c)-S \rVert^2.
    \label{eq:pred}
\end{equation}

\begin{figure*}
    \centering
    \includegraphics[width=0.4975\textwidth]{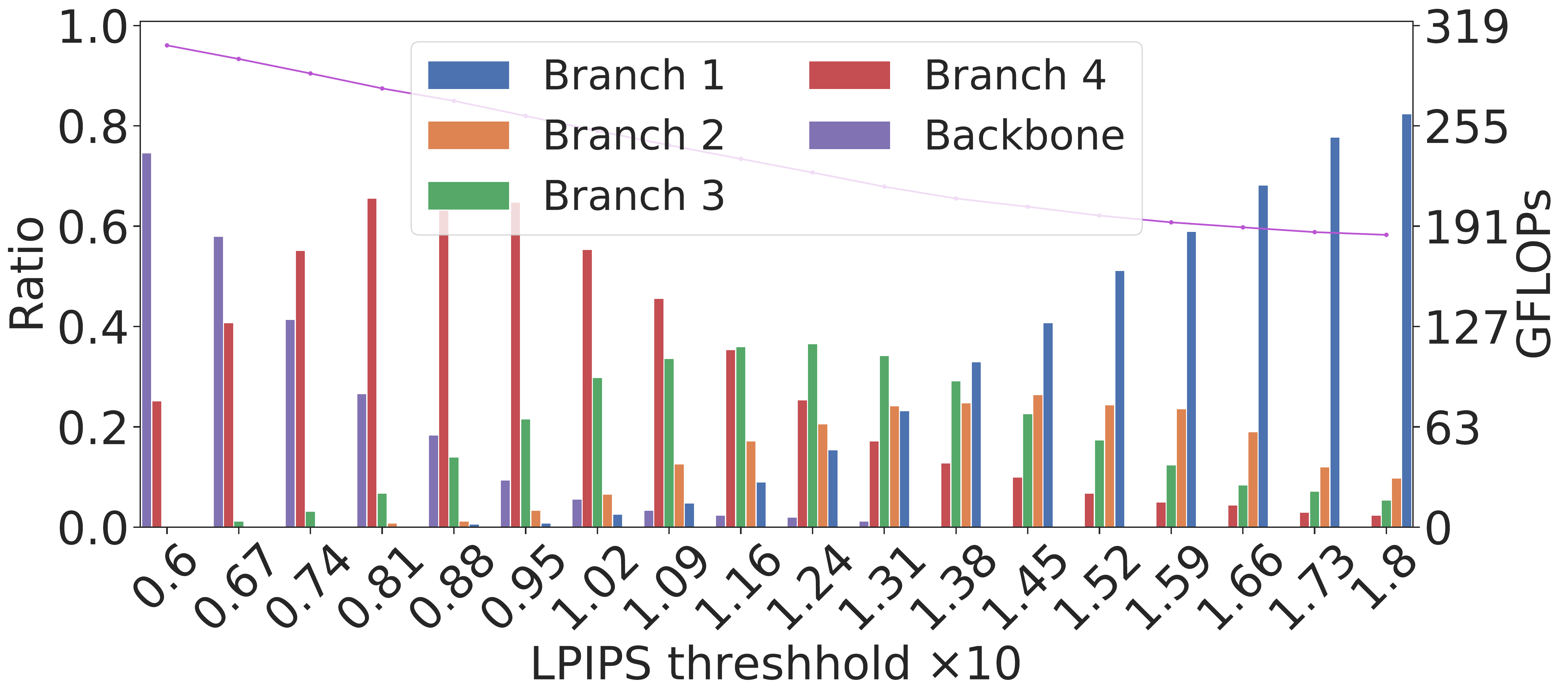}
    \includegraphics[width=0.4975\textwidth]{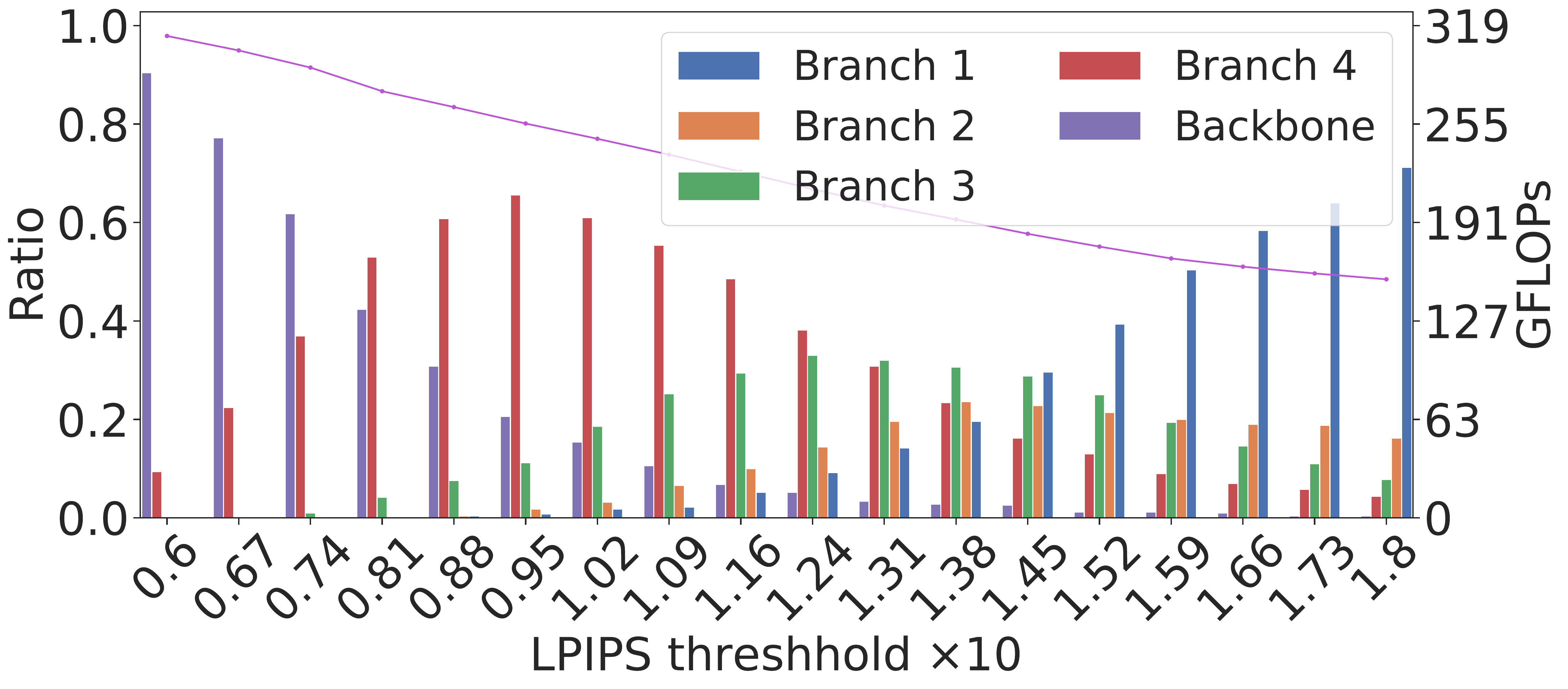}
    \includegraphics[width=0.4975\textwidth]{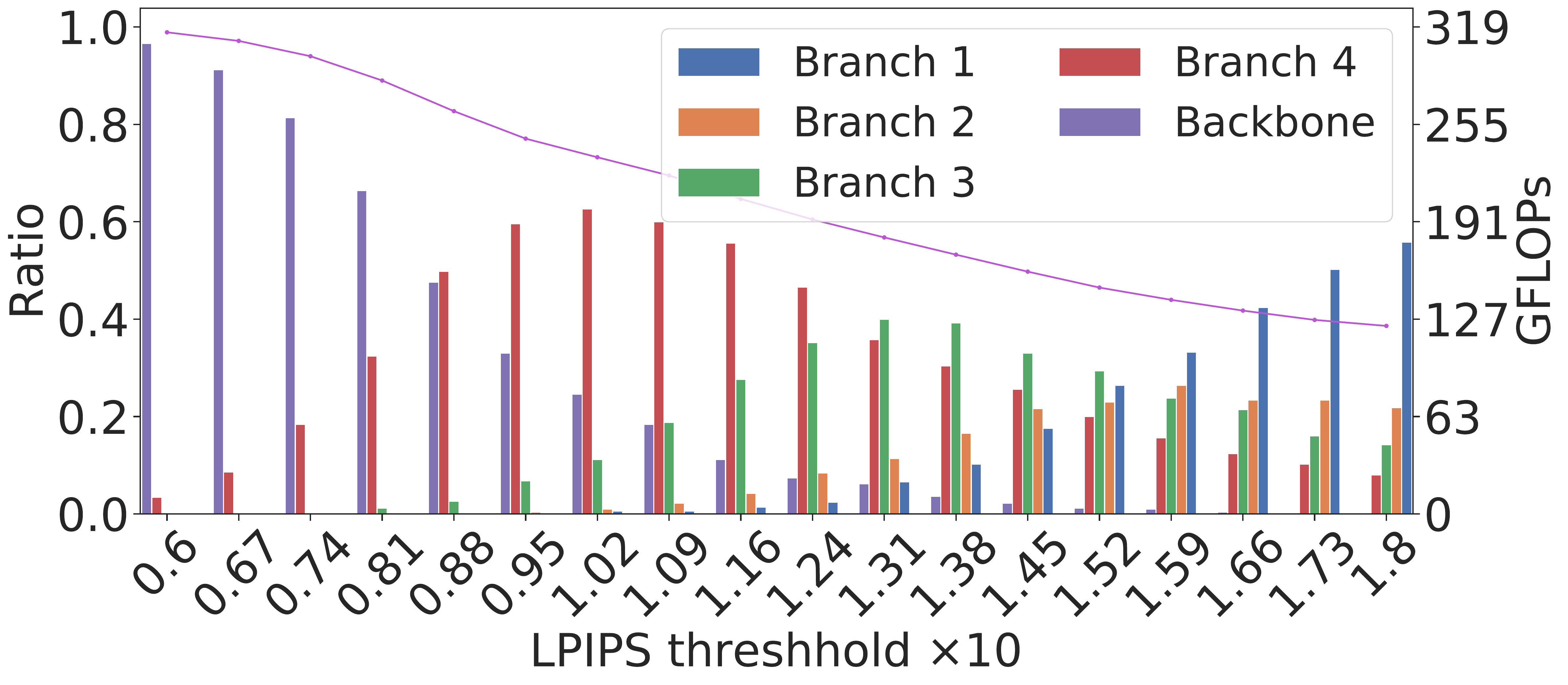}
    \includegraphics[width=0.4975\textwidth]{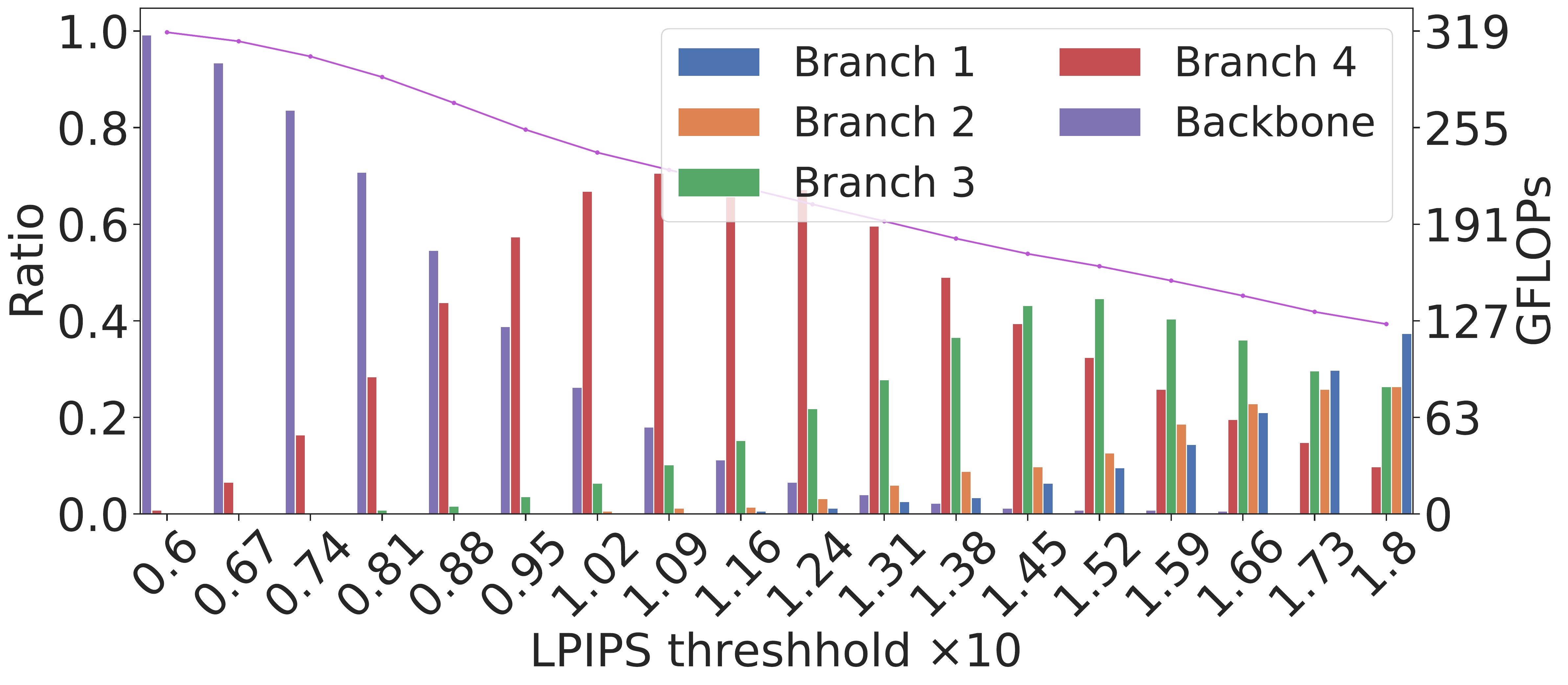}
    \caption{OASIS pipeline, min channels=32, comparison between the efficacy of different scale factors. Top to bottom, left to right: SF~$=1/2,\;1/3,\;1/4,\;1/6$.}
    \label{fig:oasis_tzar_32}
\end{figure*}



\begin{figure*}
    \centering
    \includegraphics[width=0.45\textwidth]{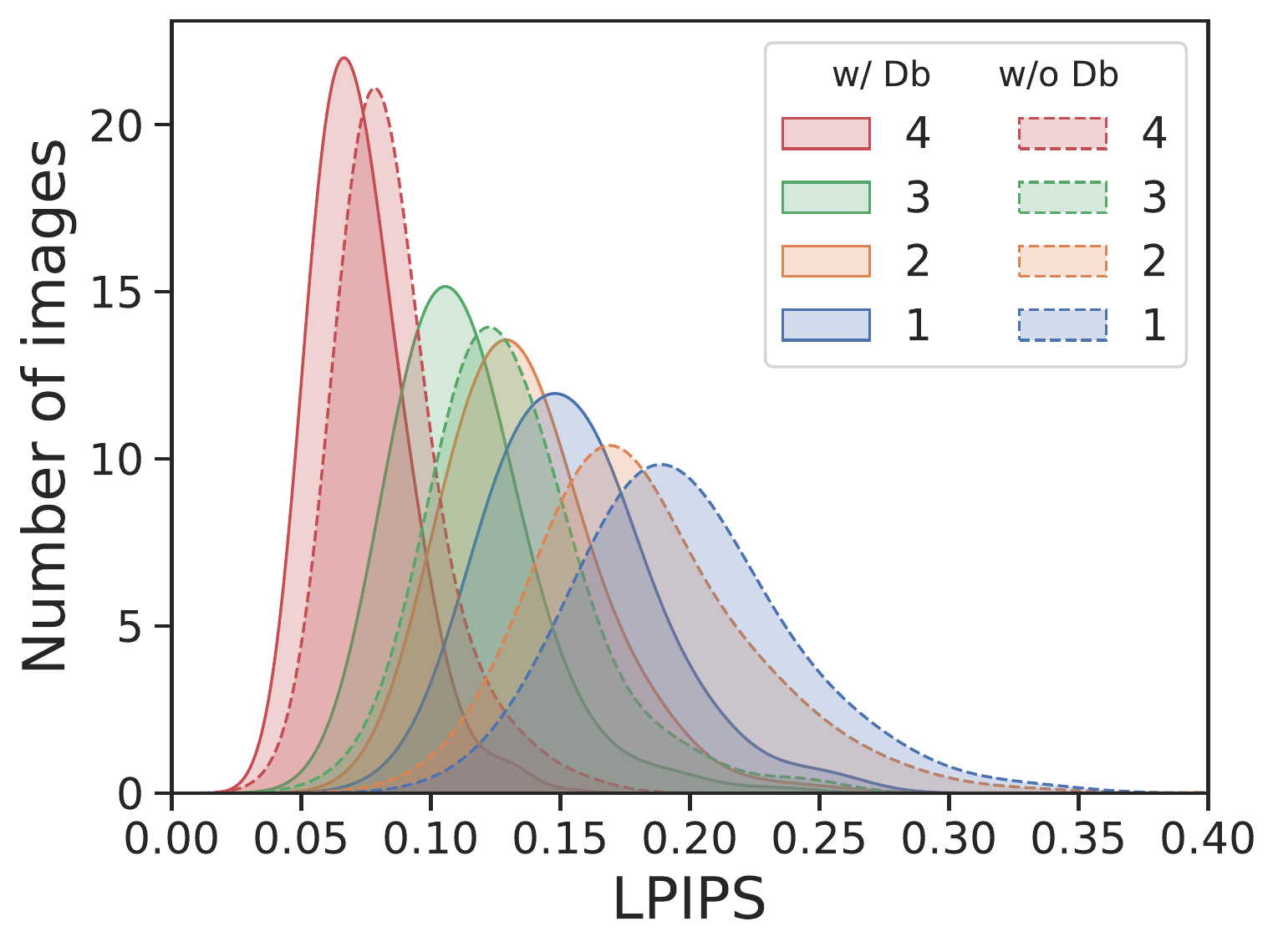}
    \includegraphics[width=0.45\textwidth]{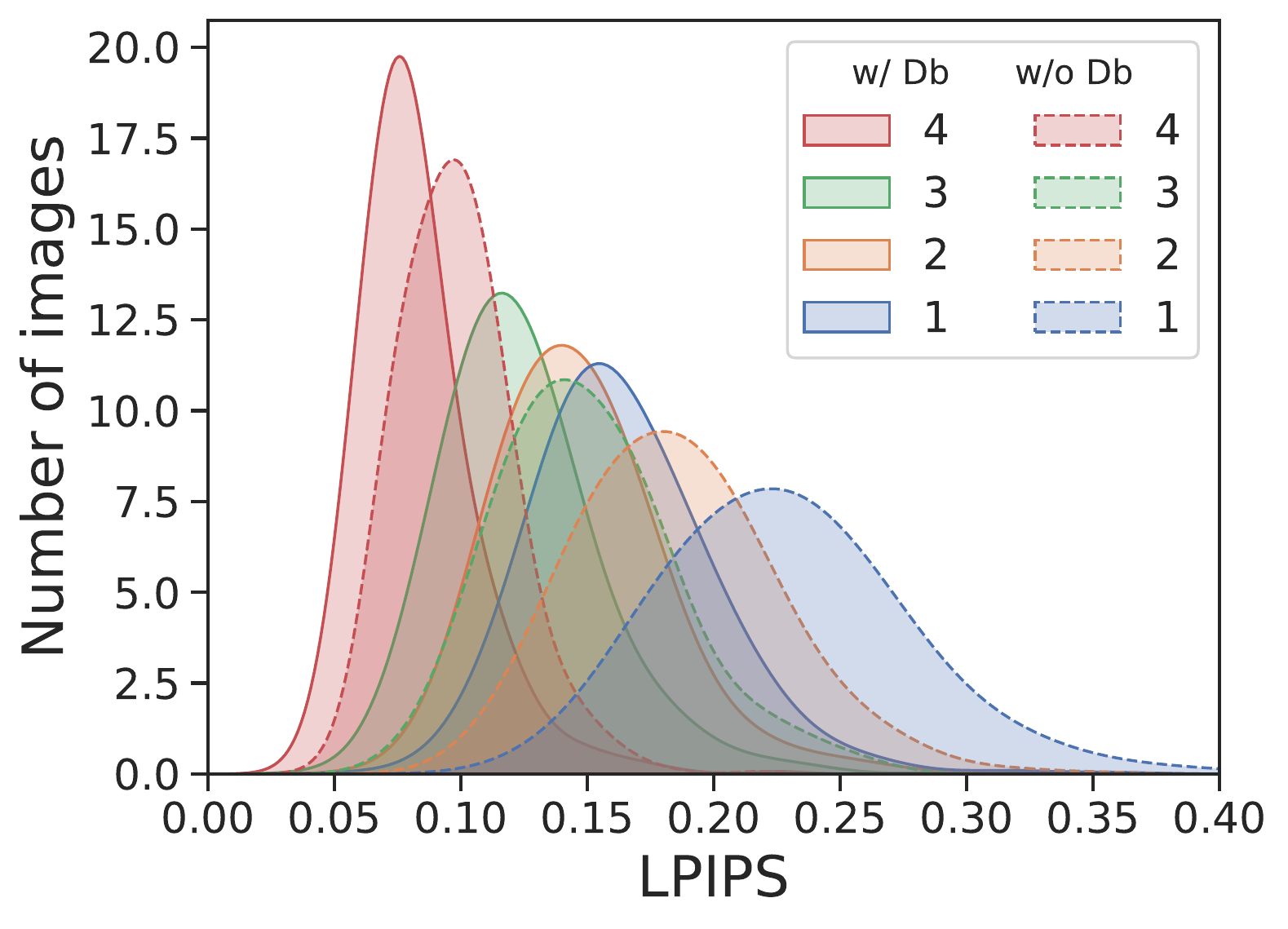}
    \includegraphics[width=0.45\textwidth]{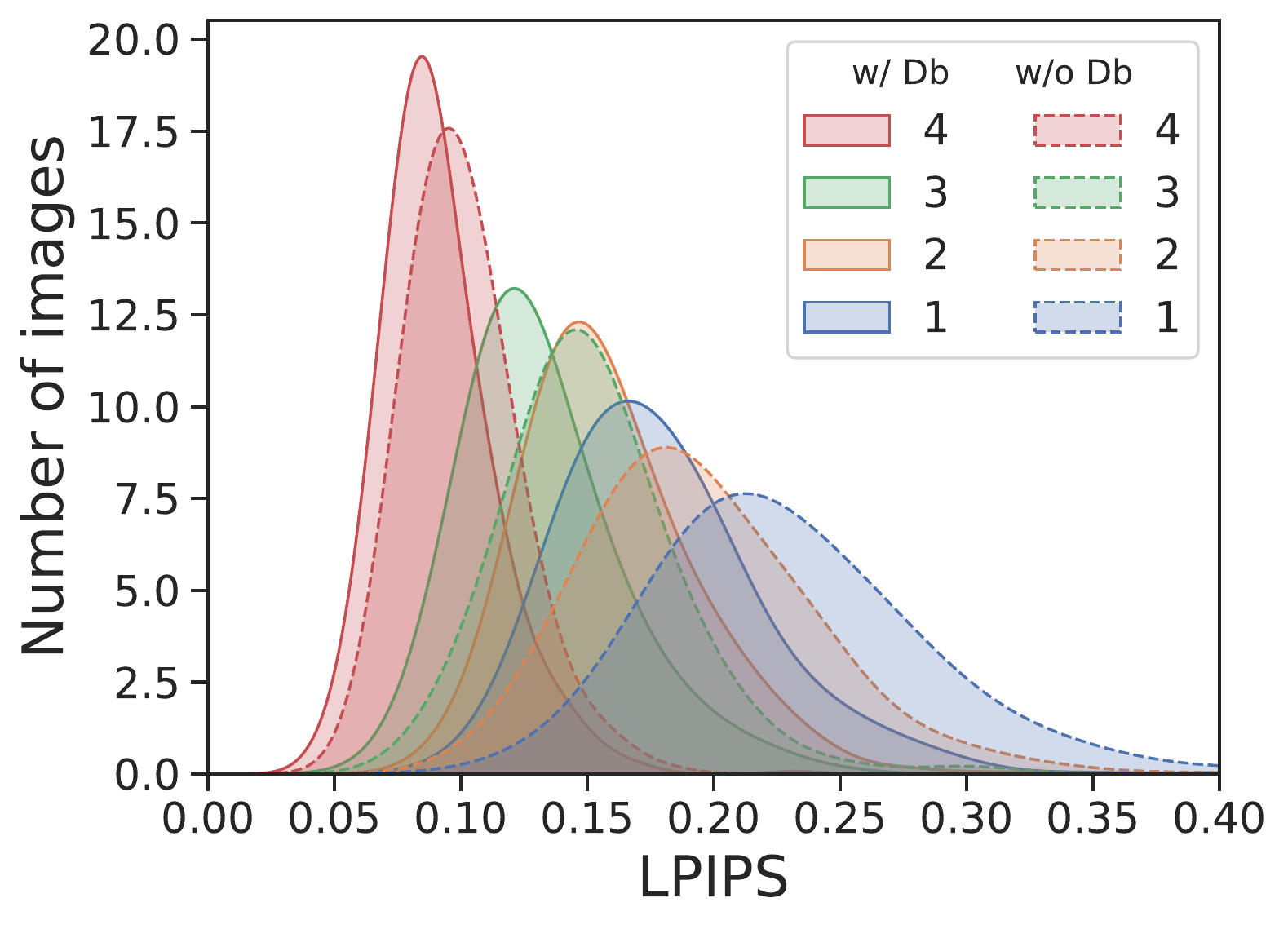}
    \includegraphics[width=0.45\textwidth]{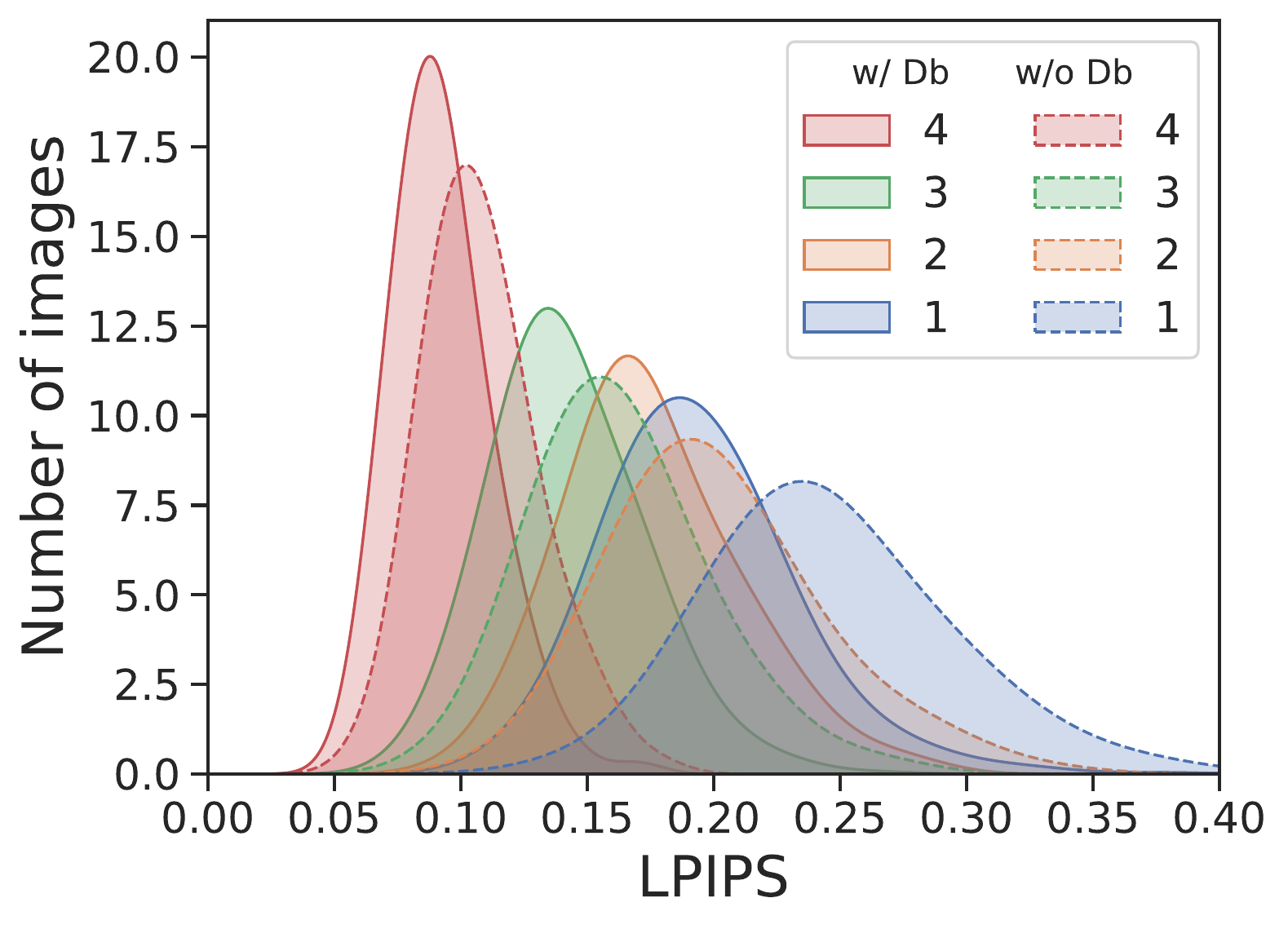}
    \caption{OASIS pipeline, comparison between database effect to quality distribution for min channels=32 and different scale factors. Top to bottom, left to right: SF~$=1/2,\;1/3,\;1/4,\;1/6$.}
    \label{fig:oasis_db_32}
\end{figure*}

\begin{table*}
    \vskip -0.1in
  \centering
  \begin{tabular}{c c |c c| c c| c c | c c}
    \toprule
    \textbf{SF} & \textbf{Bank} & \multicolumn{2}{c|}{\textbf{Branch 1}} & \multicolumn{2}{c|}{\textbf{Branch 2}} & \multicolumn{2}{c|}{\textbf{Branch 3}} & \multicolumn{2}{c}{\textbf{Branch 4}}  \\
     & & FID$\downarrow$ & mIOU$\uparrow$ & FID$\downarrow$ & mIOU$\uparrow$ & FID$\downarrow$ & mIOU$\uparrow$ & FID$\downarrow$ & mIOU$\uparrow$  \\
     
    \midrule
    \multirow{2}{*}{1/2} & \xmark & 58.8 & 62.8 & 58.3 & 63.3 & 53.2 & 66.9& 51.3 & 69.0 \\
     & \cmark & 52.3 & 65.6 & 51.4 & 67.8 & 49.6 & 67.3 & 48.6 & 67.8 \\ [0.1cm]
    \multirow{2}{*}{1/3} & \xmark & 66.8 & 57.1 & 60.7 & 62.6 & 52.8 & 65.9 & 51.9 & 66.7  \\
     & \cmark & 55.2 & 66.7 & 54.1 & 66.1 & 53.1 & 67.0 & 51.9 & 68.3   \\ [0.1cm]
    \multirow{2}{*}{1/4} & \xmark & 69.5 & 59.7 & 60.8 & 61.4 & 58.4 & 65.1 & 54.4 & 67.4  \\
      & \cmark & 57.7 & 65.9 & 57.4 & 66.7 & 55.3 & 67.5 & 51.2 & 67.7  \\ [0.1cm]
     \multirow{2}{*}{1/6} & \xmark & 69.5 & 56.7 & 65.2 & 62.1 & 61.9 & 65.1 & 54.0 & 66.4  \\
      & \cmark & 60.6 & 67.6 & 58.1 & 66.8 & 57.6 & 67.0 & 51.4 & 68.9  \\
    \midrule
    \multicolumn{2}{c}{Backbone}&
    \multicolumn{2}{c}{} & \multicolumn{2}{c}{} & \multicolumn{2}{c}{} & 47.7 & 69.3  \\
    \bottomrule
  \end{tabular}
  \caption{Quantitative results for the OASIS pipeline at different scale factors. The minimum number of channels is 32.}
  \label{tab:metrics_oasis_fid_miou_min32}
\end{table*}

The learning rate was set to $0.01$, the loss was optimized via stochastic gradient descent with cosine scheduler \cite{loshchilov2017cosine_warm}.
The choice of training set for the predictor was not trivial, since the pipeline inputs consist of a semantic map concatenated to a 3D noise tensor. Due to the high dimensionality of the noise space, sampling uniformly from it does not guarantee any convergence for the learning process. Instead, we randomly extracted $100$ 3D noise tensors and combined them with $500$ semantic maps from the Cityscapes~\cite{Cordts2016cityscape} training set, thus obtaining $50\;000$ examples. We then tested this technique by using $300$ and $500$ noise tensors. Once trained, we measured the predictor's error by using $500$ images from Cityscapes' validation set combined with the same noises used for the training and with new noises. The results are reported, respectively, in \Cref{tab:OASIS_pred_val} and \Cref{tab:OASIS_pred_test}.

\begin{table}[H]
    \centering
    \begin{tabular}{c c c c c c}
    \toprule
         Noises&B 1&B 2&B 3&B 4&Mean error\\
    \midrule
         100 & 5\% & 6\% & 6\% & 7\% & 6\%\\
         300 & 5\% & 5\% & 6\% & 6\% & 5.5\%\\
         500 & 5\% & 5\% & 6\% & 6\% & 5.5\%\\
    \bottomrule
    \end{tabular}
    \caption{Validation error for the OASIS predictor. The validation set was created joining the noises used for the training to the $500$ semantic maps from the validation set of the Cityscapes dataset.}
    \label{tab:OASIS_pred_val}
    \vskip -0.2in
\end{table}

\begin{table}[H]
    \centering
    \begin{tabular}{c c c c c c}
    \toprule
         Noises&B 1&B 2&B 3&B 4&Mean error\\
    \midrule
         100 & 14\% & 14\% & 13\% & 16\% & 14\%\\
         300 & 10\% & 11\% & 11\% & 15\% & 12\%\\
         500 & 10\% & 10\% & 10\% & 13\% & 11\%\\
    \bottomrule
    \end{tabular}
    \caption{Test error for the OASIS predictor. The test set was created joining random noises to the $500$ semantic maps from the validation set of the Cityscapes dataset.}
    \label{tab:OASIS_pred_test}
\end{table}

\subsection{The MegaPortraits pipeline}
\subsubsection{Architectures and dimensions}

  
    
 

\begin{table}
  \centering
  \textbf{Cross-reenactment}
  
   \begin{tabular}{c c | c|  c|  c }
    \toprule
    SF & Bank & Branch 1 & Branch 2 & Branch 3   \\
     & & FID$\downarrow$ & FID$\downarrow$ & FID$\downarrow$  \\
     
    \midrule
    \multirow{2}{*}{1/3} & \xmark & 56.05 & 52.77 & 49.08 \\
     & \cmark & 54.60 & 52.40 & 50.44  \\ [0.1cm]
    \multirow{2}{*}{1/6} & \xmark & 61.30 & 55.58 & 51.00 \\ 
     & \cmark & 59.01 & 54.08 & 50.84  \\ [0.1cm]
    \multirow{2}{*}{1/8} & \xmark & 61.84 & 55.66 & 50.88 \\
     & \cmark & 57.94 & 54.88 & 50.96  \\ [0.1cm]
     \multirow{2}{*}{1/15} & \xmark & 66.87 & 61.75 & 51.56 \\
     & \cmark & 57.25 & 57.70 & 51.85  \\
    \midrule
    \multicolumn{2}{c}{Backbone}&
      \multicolumn{1}{c}{} &  \multicolumn{1}{c}{} & 50.28  \\
    \bottomrule
  \end{tabular}
 
  \caption{Quantitative results for the MegaPortraits pipeline, cross-reenactment.}
  \label{tab:metrics_avatar_cross}
  \vskip -0.1in
\end{table}

The original MegaPortraits \cite{Drobyshev22megaportraits} generative DNN for images of resolution $512\times512$ pixels consists of a set of modules predicting a volumetric representation and another set, called G2D, that renders an output image from a processed volume. Its total number of parameters is 32M. We appended our branches after ResBlock2D modules $2,\;4,\;6$. Their respective length is $7,\;5,\;3$. Just as before, we created lighter computational paths by scaling down all channels uniformly. The new channel numbers were obtained multiplying the original ones by a scale factor. As before, we restricted the effect of this scaling by imposing a minimum number of channels equal to 24, under which no further scaling was forced. We enforced a plethora of different scale factors,
which we report in \Cref{tab:avatars_ch18}.
For this task, we used a database containing $960$ key-value pairs. The values consisted of RGB images of the source subject, uniformly covering the space of head rotations and expressions. The keys were obtained exploiting the MegaPortraits initial modules, the so-called encoders, that yield the Euler angles at which a head is rotated, as well as a multitude of parameters encoding face expressions.
Each key encoded $3$ angles and a $512$-dimensional vector for the expressions. The total size of stored parameters is therefore $0.9$G.

The database was searched for the closest key during the inference phase with the aid of the FAISS library~\cite{johnson2019faiss}. Each retrieved image was subsequently concatenated to the input of all ResBlock2D modules in every branch, thus when employing the database $3$ channels must be added to all input channels in \Cref{tab:avatars_ch18}. The architecture of the MegaPortraits predictor is summarized in \Cref{tab:avatar_pred}.

\begin{table}
    \centering
    \begin{tabular}{ c c c c c}
    \toprule
         B 1&B 2&B 3&Mean error\\
    \midrule
          1\% & 1\% & 2\% & 1\%\\

    \bottomrule
    \end{tabular}
    \caption{Test error for the MegaPortraits predictor for SF = 1/8.}
    \label{tab:avatars_pred_val}
\end{table}

\begin{figure}
    \centering
    \includegraphics[width=0.45\textwidth]{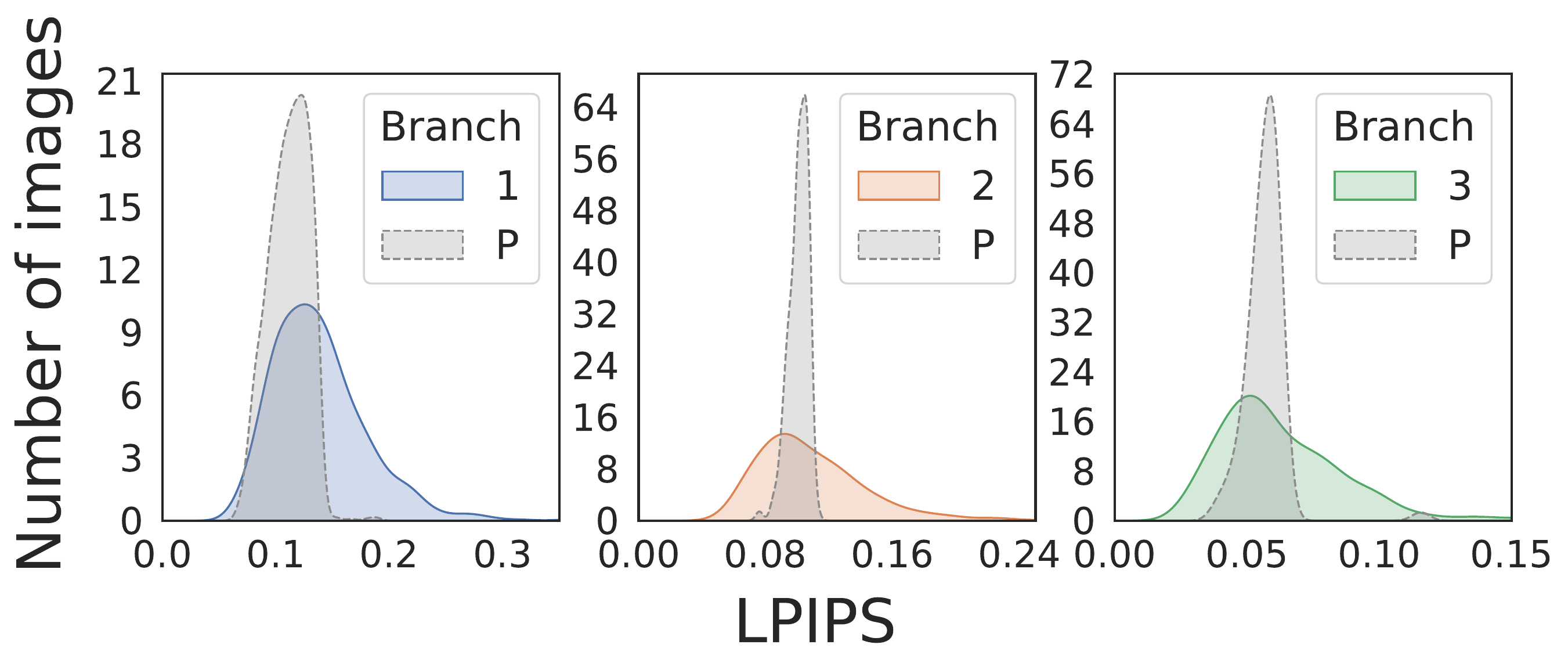}
    \caption{Comparison between quality distributions of single MegaPortraits branches, and quality distributions obtained by use of the predictor (P). The predictor was set to enforce thresholds equal to the branches’ mean quality. LPIPS were obtained by comparing images of branches for SF=1/8 with backbones’ images. The curves are the result of kernel density estimation with
bandwidth 0.3.}
    \label{fig:avatar_density}
\end{figure}

\subsubsection{Training details}
\begin{figure*}
    \centering
    \includegraphics[width=0.4975\textwidth]{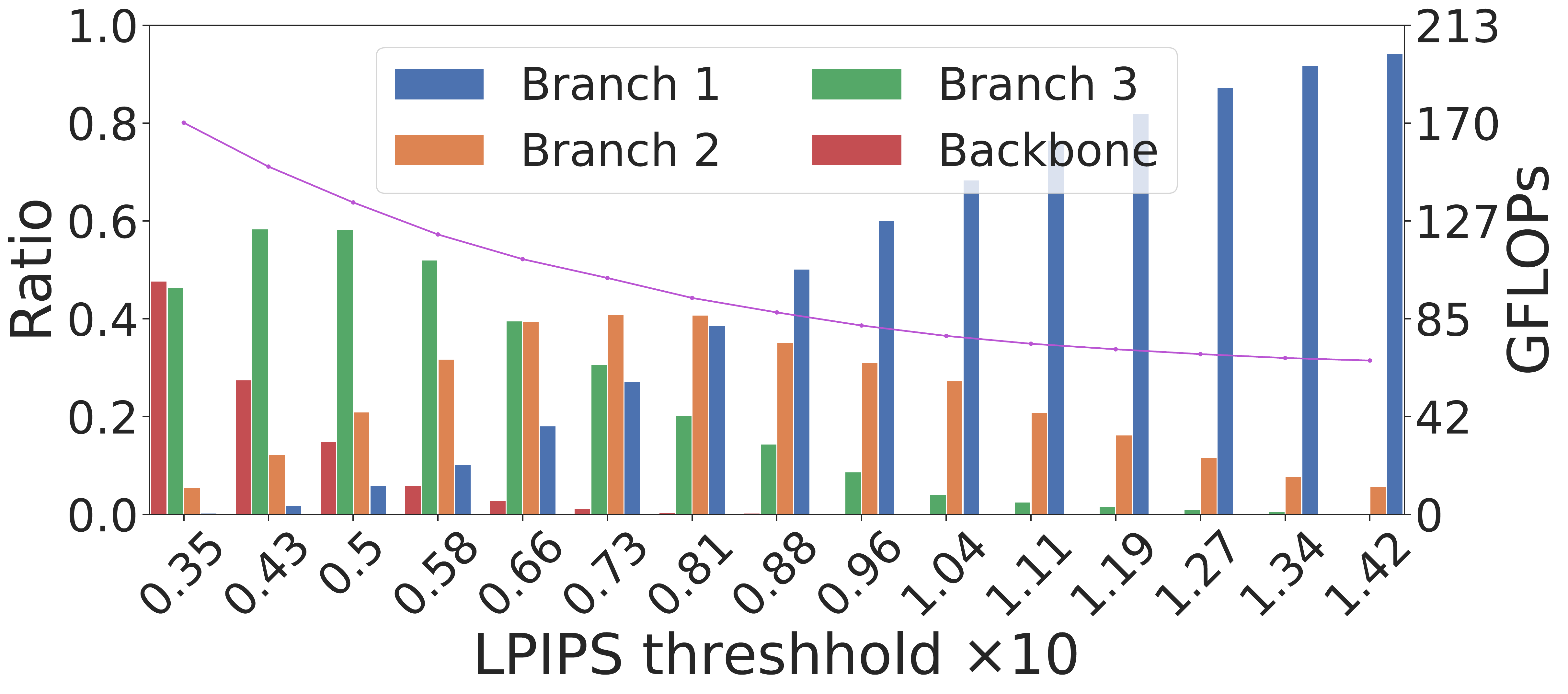}
    \includegraphics[width=0.4975\textwidth]{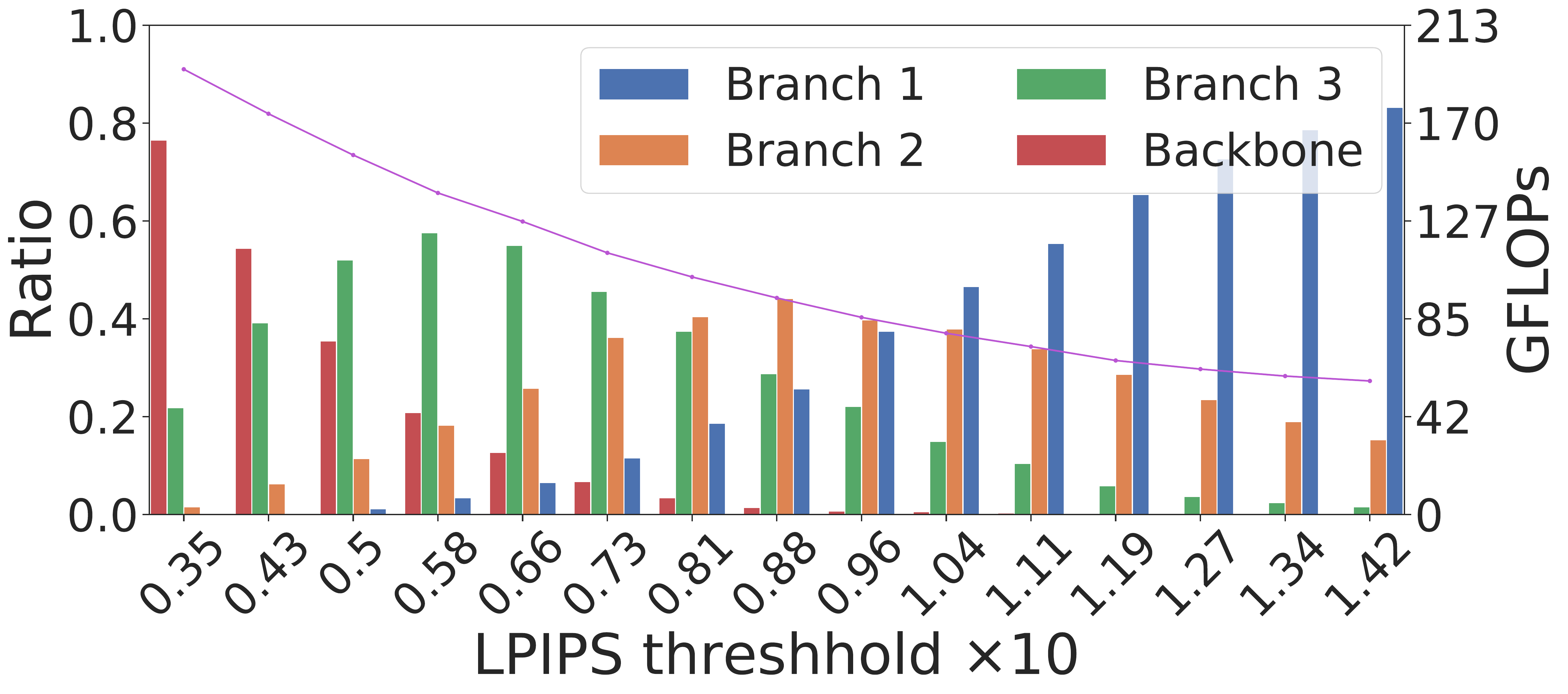}
    \includegraphics[width=0.4975\textwidth]{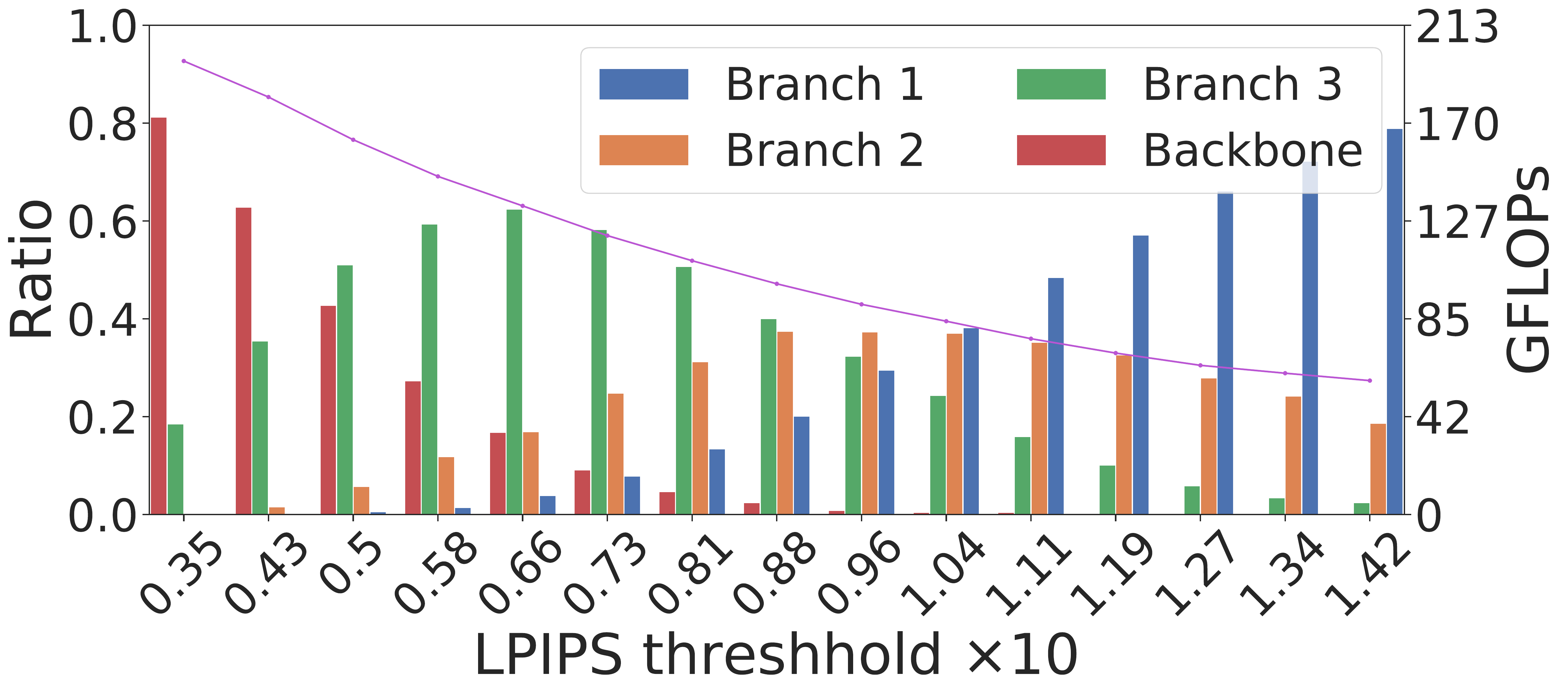}
    \includegraphics[width=0.4975\textwidth]{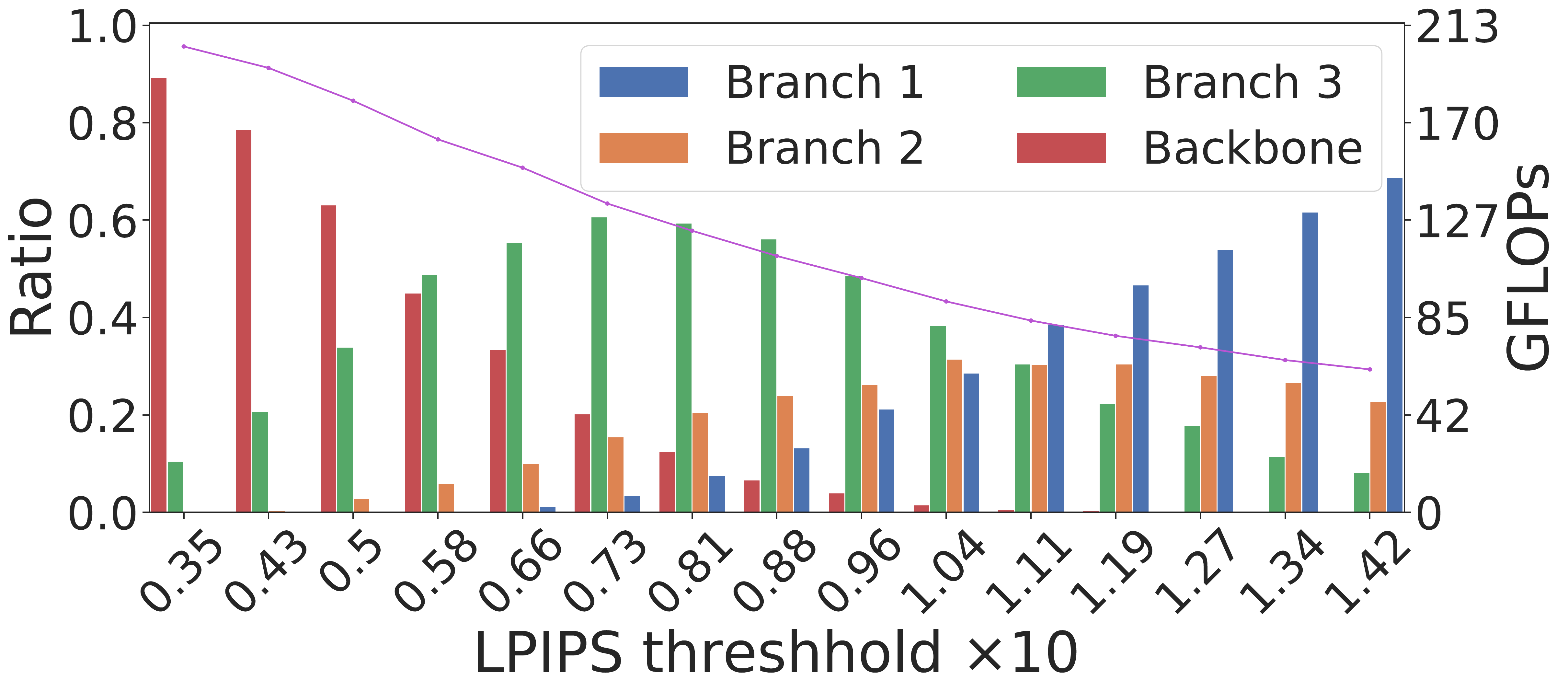}
    \caption{MegaPortraits pipeline, comparison between the efficacy of different scale factors. From left to right: SF~$=1/3,\;1/6,\;1/8,\;1/15$.}
    \label{fig:avatar_tzar}
\end{figure*}

\begin{figure*}
    \centering
    \includegraphics[width=0.4\textwidth]{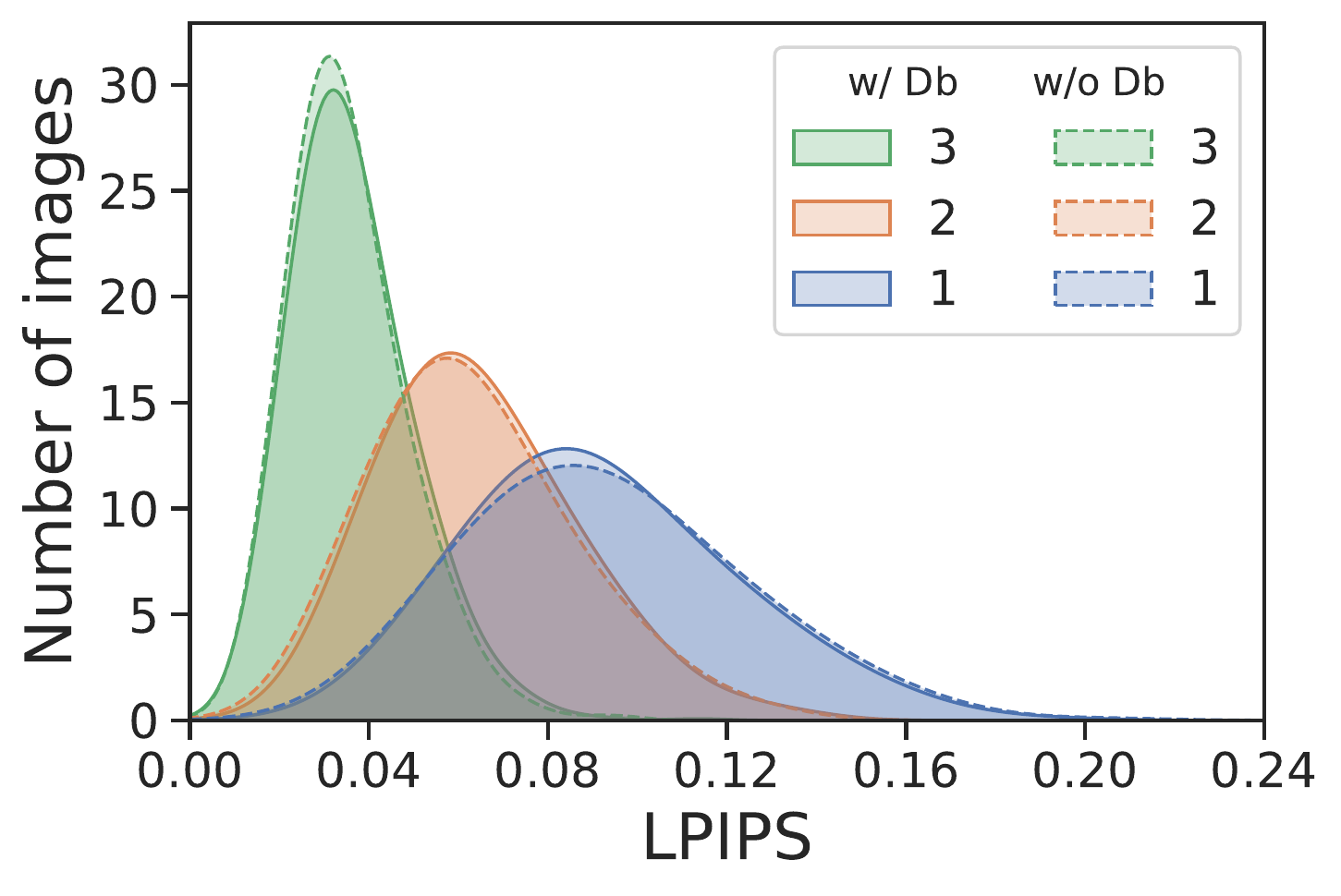}
    \includegraphics[width=0.4\textwidth]{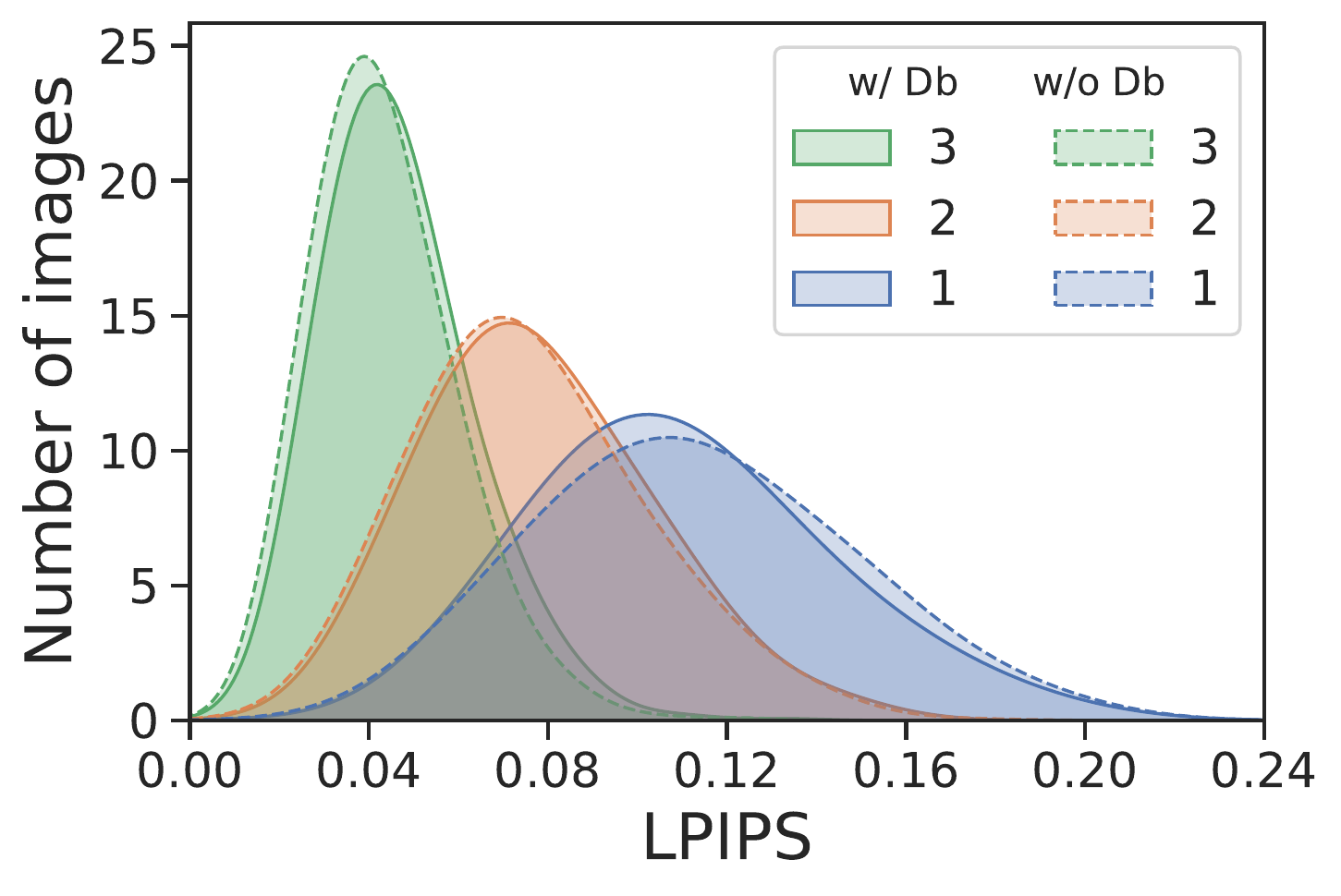}
    \includegraphics[width=0.4\textwidth]{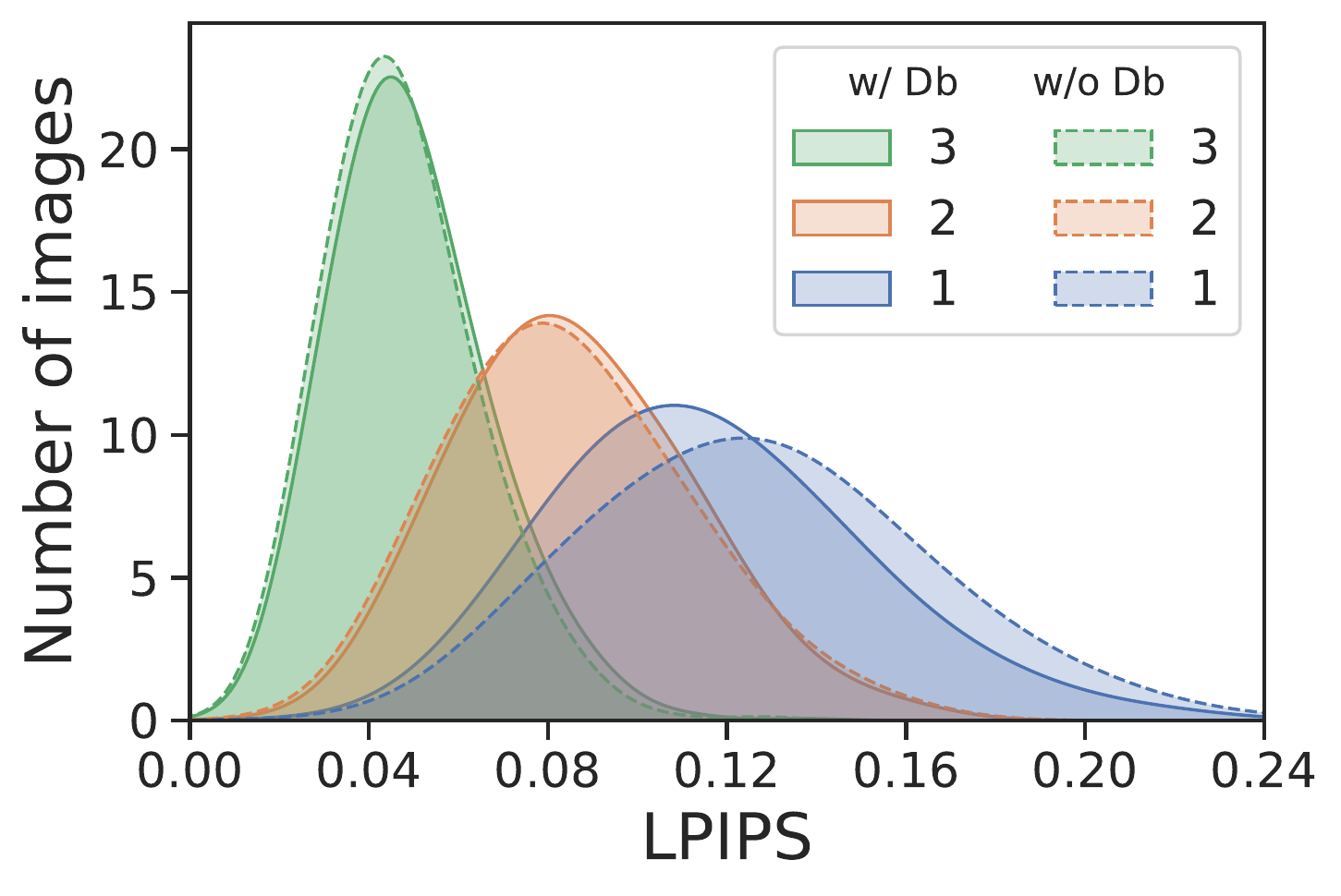}
    \includegraphics[width=0.4\textwidth]{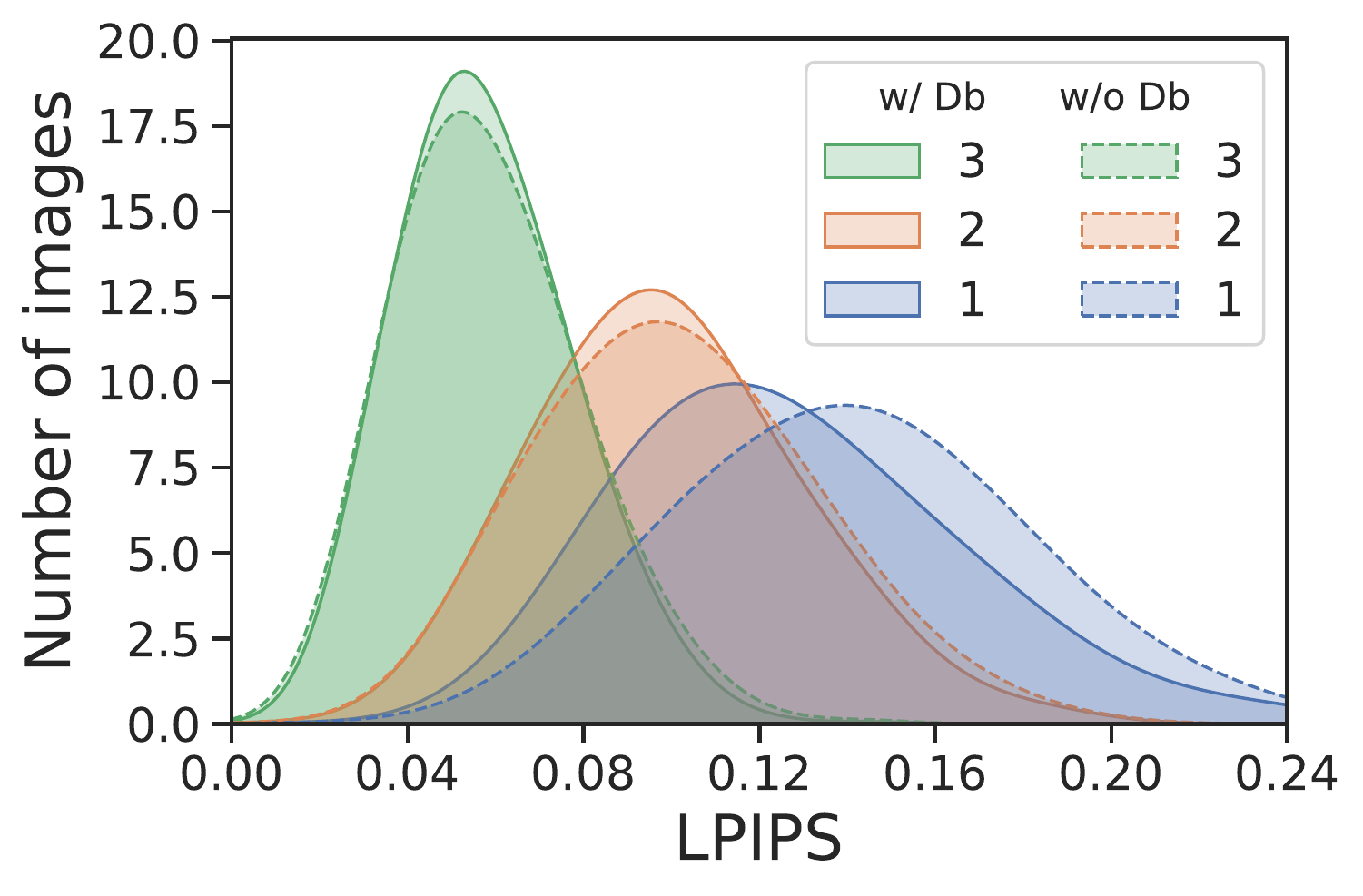}
    \caption{MegaPortraits pipeline, comparison between database effect to quality distribution for different scale factors. Left to right: SF~$=1/3,\;1/6,\;1/8,\;1/15$.}
    \label{fig:avatar_db}
\end{figure*}

For the MegaPortraits pipeline, we trained our branches using hinge adversarial loss, each branch competing against a copy of multi-scale patch discriminator \cite{zhu2017cycleGAN}. Additionally, we imposed feature matching \cite{wang2018feature_matching}, VGG19 perceptual \cite{johnson2016vgg}, L1 and MS-SSIM \cite{wang2003mssim} losses. We also use a specialized gaze loss computed with a VGG16 network that distills gaze detection (RT-GENE, \cite{FischerECCV2018}) and blink detection (RT-BENE, \cite{CortaceroICCV2019W}) systems into one model. More details on the losses can be found in MegaPortraits \cite{Drobyshev22megaportraits}. All losses are computed in relation to the backbone images and using only foreground regions. Overall, the total loss is
\begin{multline}
\mathcal{L}_{\text{Branch}}= \mathcal{L}_{\text{Adv}} + c_1 \mathcal{L}_{\text{VGG}} +  c_2 \mathcal{L}_{\text{MS-SSIM}} + c_3 \mathcal{L}_{\text{L1}} + \\ c_4 \mathcal{L}_{\text{FM}} + c_5 \mathcal{L}_{\text{GL}}
\end{multline}
with the following weights: $c_1=18$, $c_2=0.84$, $c_3=0.16$, $c_4=40$, and $c_5=5$. Branches and discriminators were trained using AdamW optimizers \cite{loshchilov2018adamw} with $\beta_1 = 0.05$, $\beta_2=0.999$, $\epsilon=10^{-8}$, weight decay $=10^{-2}$ and initial learning rate $=2\times10^{-4}$.  Cosine learning rate schedulers were employed during training with minimum learning rate of $10^{-6}$. Computations were done via PyTorch distributed data parallel. The model was trained in mixed precision on 2 P40 NVIDIA GPUs with effective batch size 6 for approximately 3 days. The resultant qualities can be found in \cref{tab:metrics_avatar_cross}.

For each input, the Predictor estimates LPIPS for all branches. To train it, we imposed MAE loss between predicted and state of truth similarity: $\mathcal{L}_{\text{Pred}}(z,c;S)= | P(z,c)-S |$.
We employed the AdamW optimizer with $\beta_1 = 0.05$, $\beta_2=0.999$ and initial learning rate $2\times10^{-4}$ alongside cosine learning rate scheduler.



\section{Comparisons}

We implemented all architectures listed in \Cref{tab:OASIS_m64} and \Cref{tab:avatars_ch18}. The overall results for the OASIS pipeline can be compared in \cref{fig:OASIS_tzars} and \cref{fig:oasis_tzar_32}, while for the MegaPortraits pipeline they are shown in \cref{fig:avatar_tzar} . We can see how different scale factors yield different branch distributions. The effect of the database on the branches of all scale factors is reported in \cref{fig:OASIS_db}.

\section{Complexity analysis}
\begin{figure}[H]
    \centering
    \includegraphics[width=0.4\textwidth]{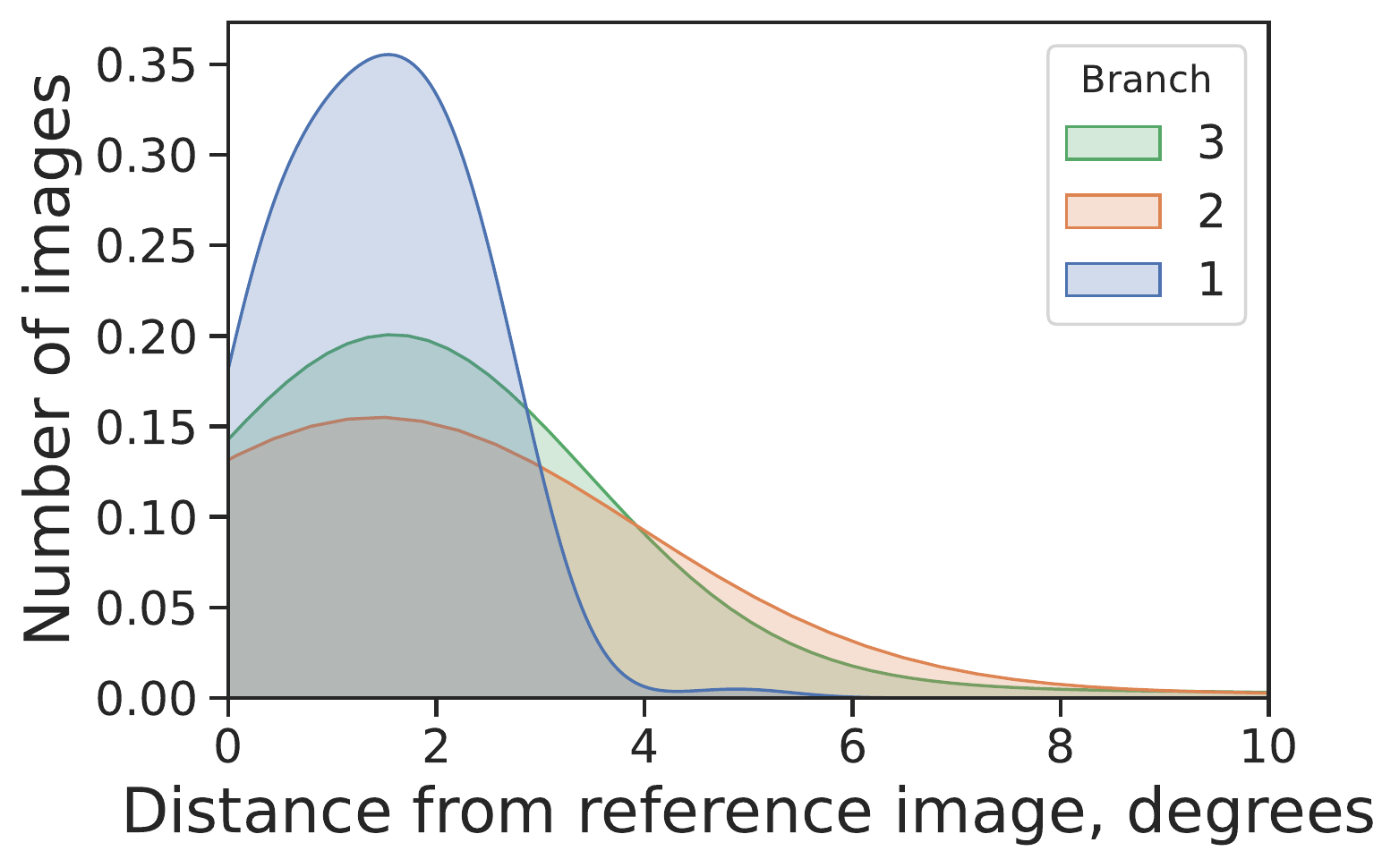}
    \includegraphics[width=0.4\textwidth]{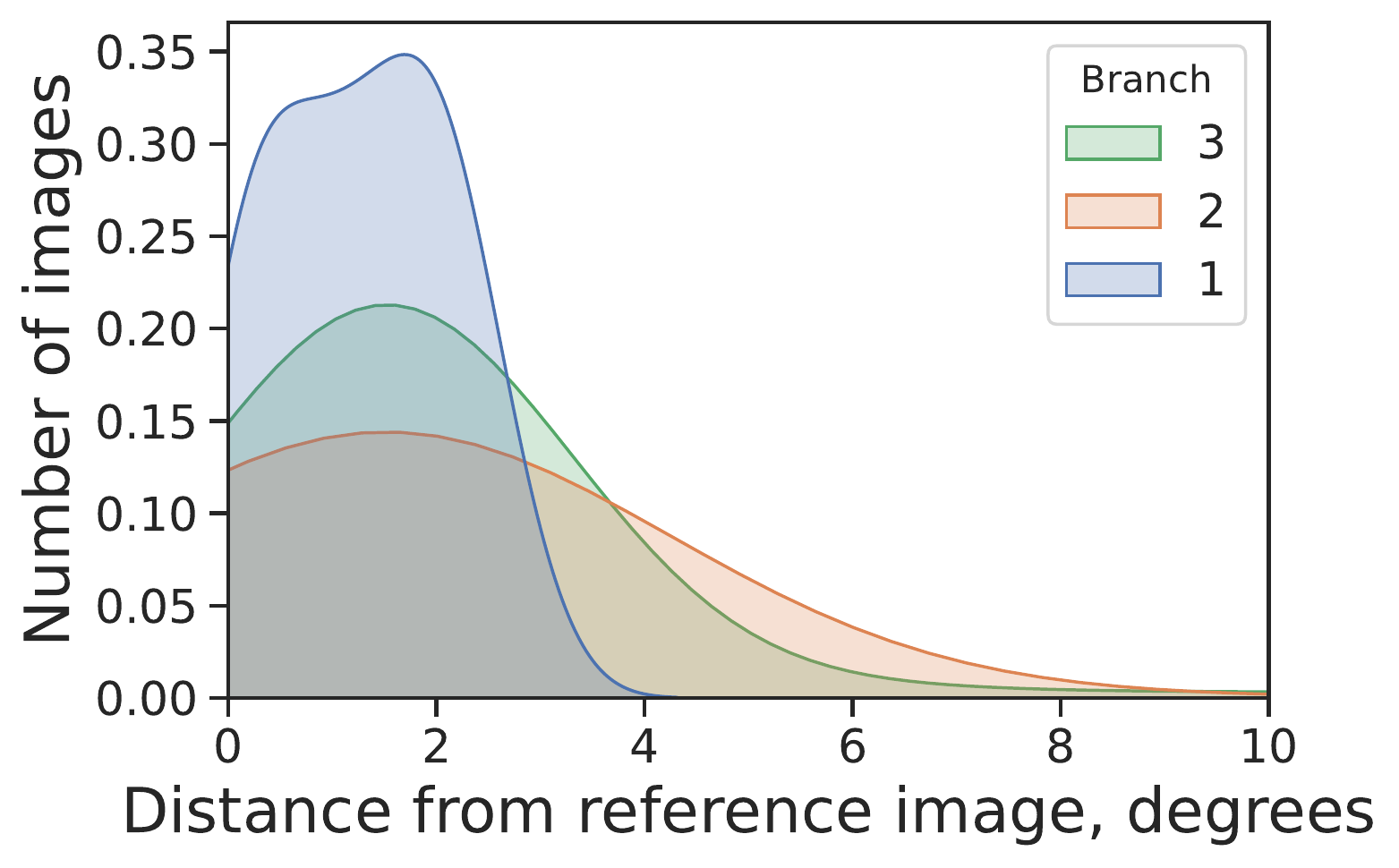}
    \caption{MegaPortraits pipeline, distribution of images routed to different branches in relation to their head rotation angle. First row SF~$=1/8$, second row SF~$=1/15$.}
    \label{fig:avatars_distance}
\end{figure}

For the MegaPortraits pipeline, the quality of synthesized images seems to correlate with the angle at which the head is rotated. This is reflected in our method as well. Indeed, heads rotated at higher angles have greater probability of being routed to a later branch, as evidenced by \cref{fig:avatars_distance}.

\begin{figure*}[p]
    \centering
    \includegraphics[width=\textwidth]{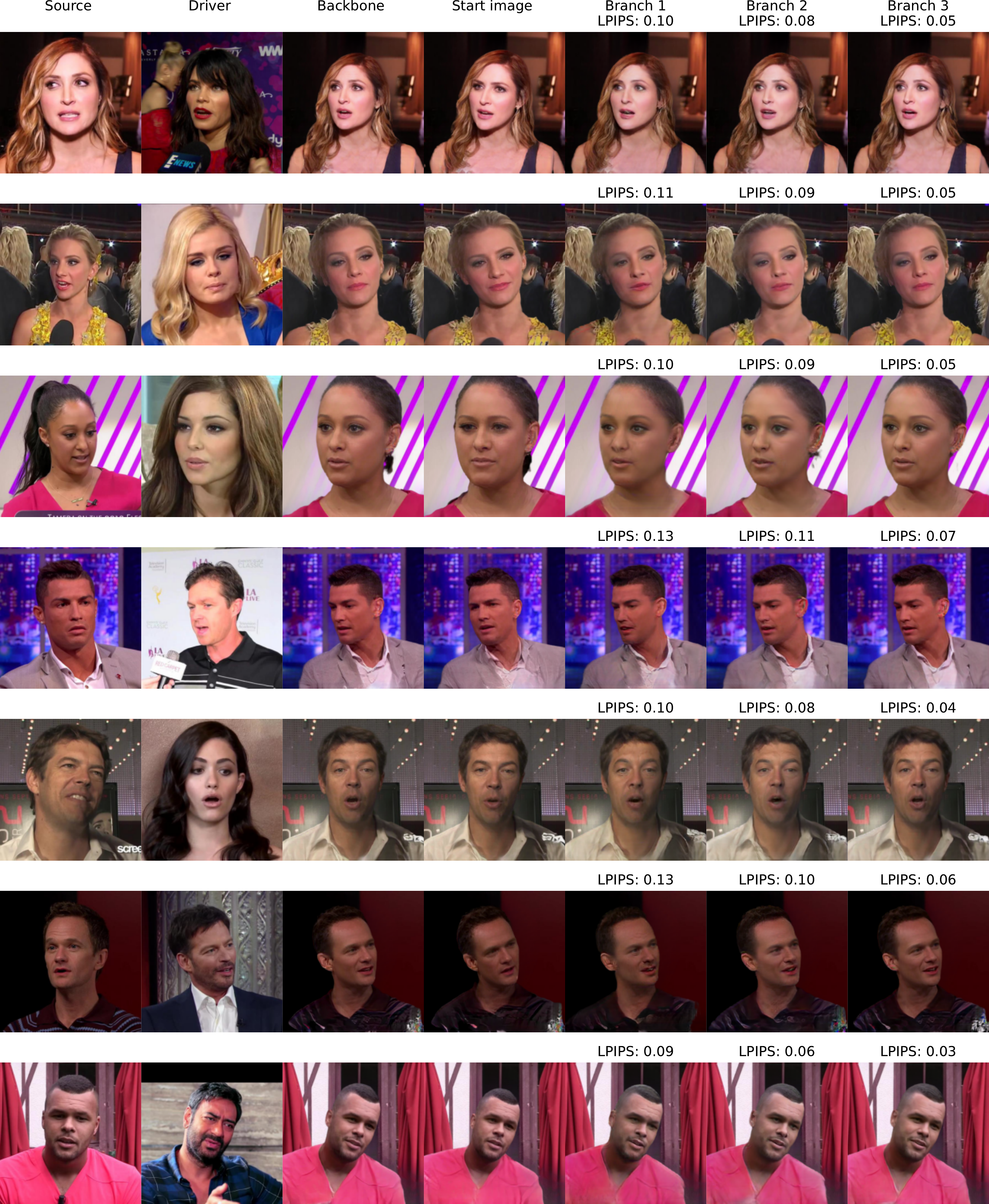}
    \caption{Samples for the MegaPortraits pipeline on SF = 1/8. The background is inpainted.}
    \label{fig:sample_avatars}
\end{figure*}

\begin{figure*}
    \centering
    \includegraphics[width=0.99\textwidth]{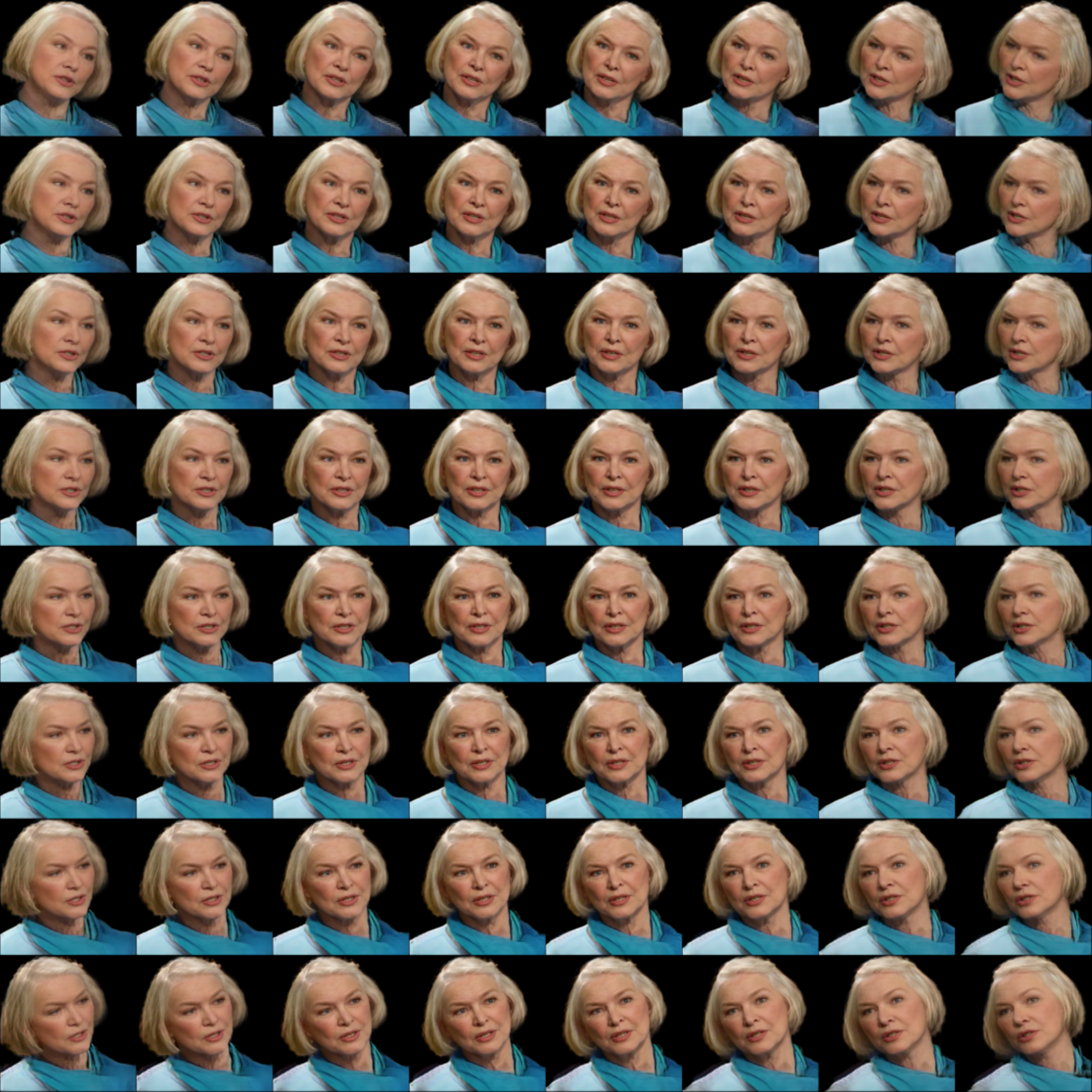}
    \vskip +0.2in
    \includegraphics[width=0.99\textwidth]{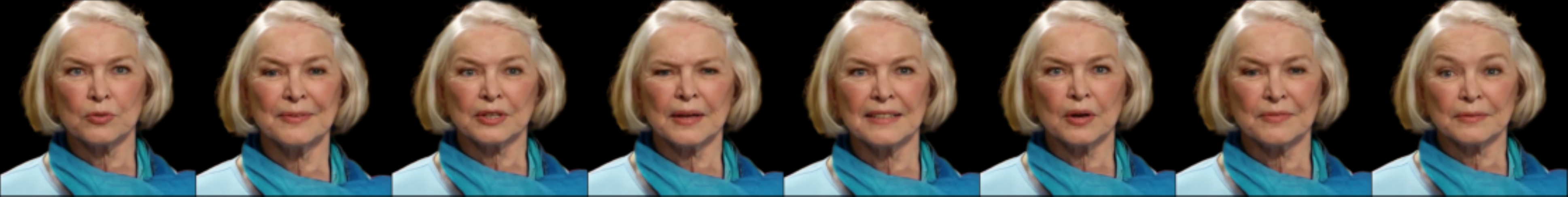}
    \caption{Examples of bank images for different rotations (top) and expressions (bottom).}
    \label{fig:avatar_db}
\end{figure*}

\begin{figure*}[p]
    \centering
    \includegraphics[width=\textwidth]{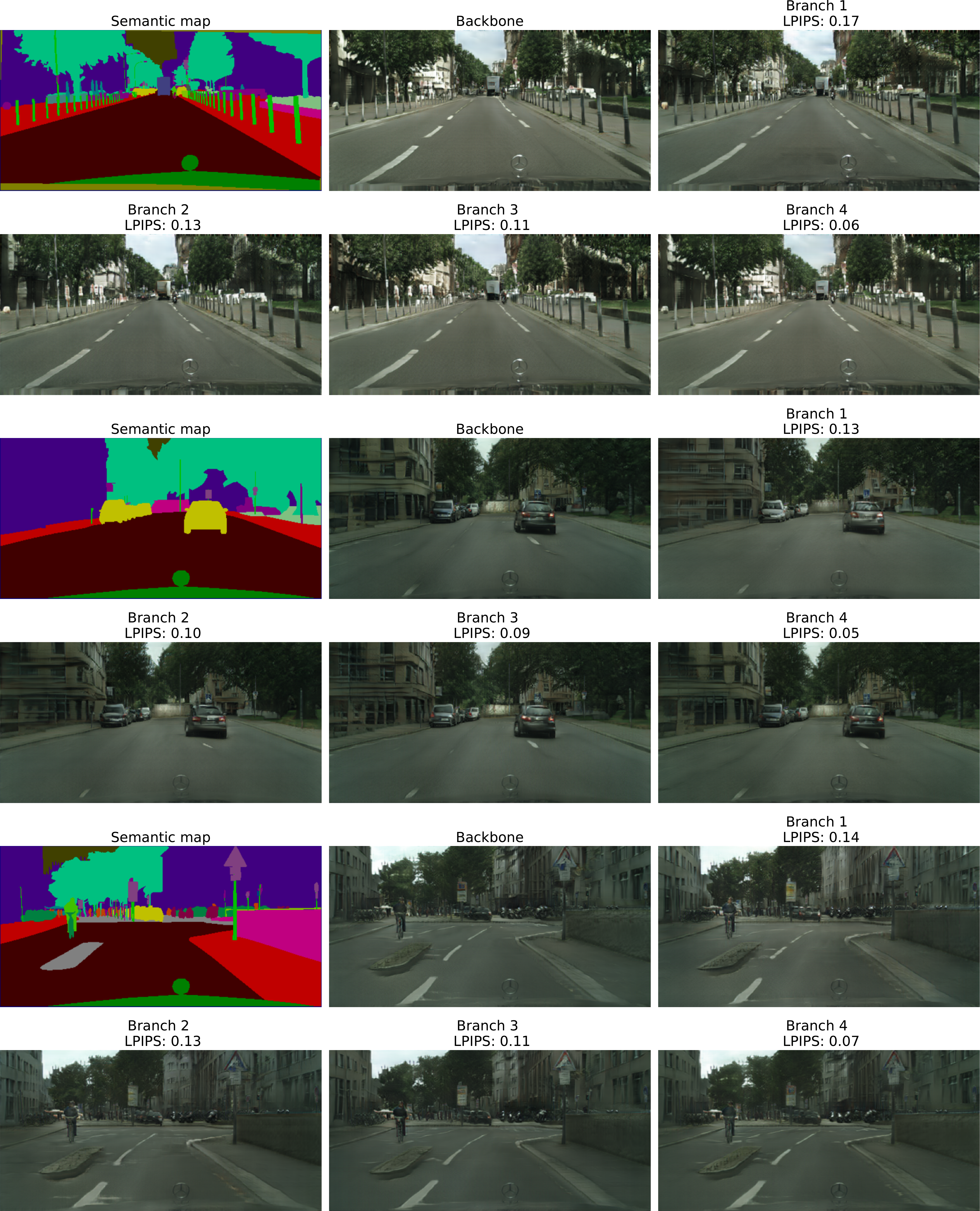}
    \caption{Samples for the OASIS pipeline on SF = 4}
    \label{fig:sample_oasis}
\end{figure*}

\begin{figure*}[p]
    \centering
    \includegraphics[width=\textwidth]{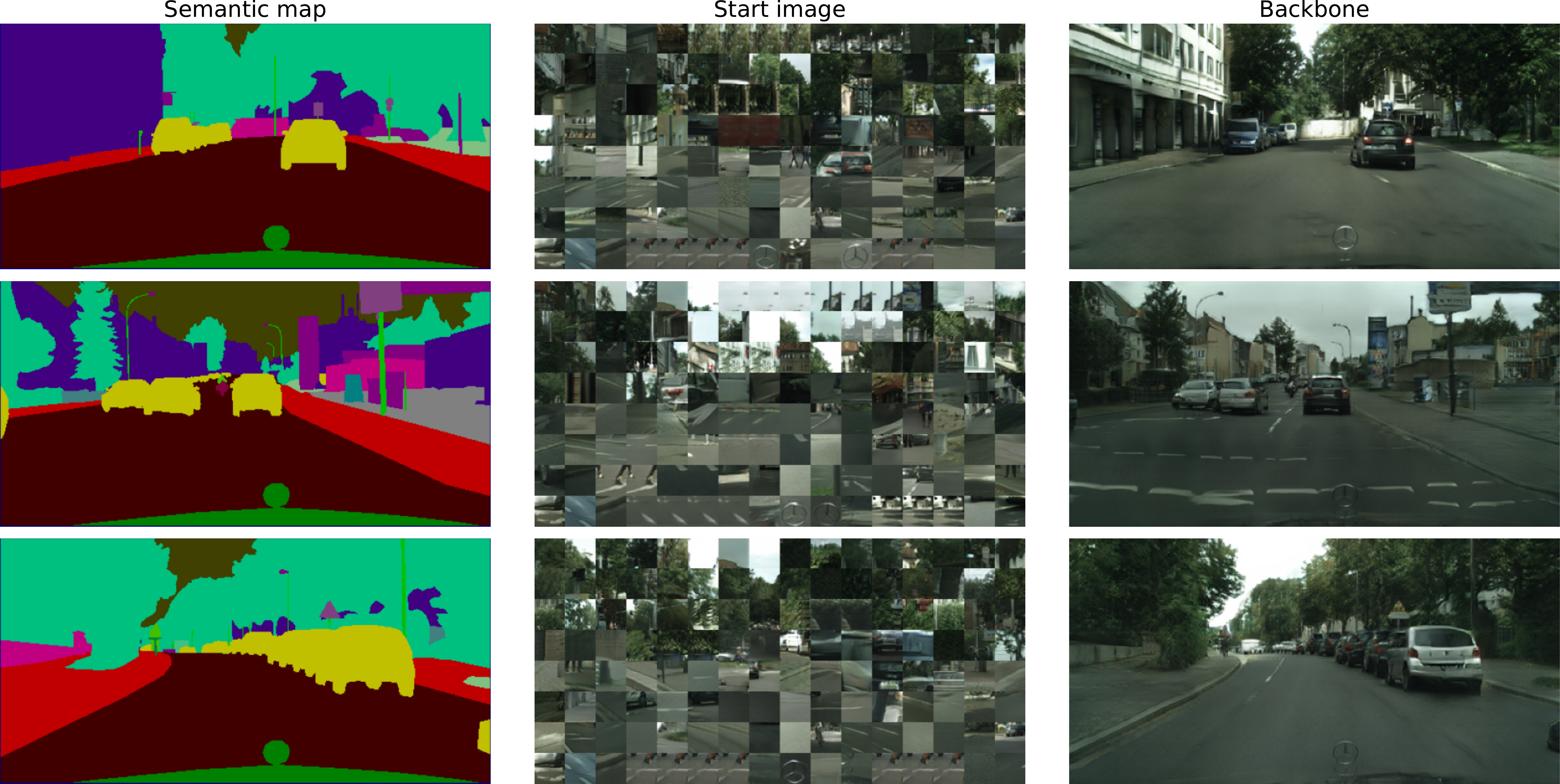}
    \vskip +0.05in
    \includegraphics[width=\textwidth]{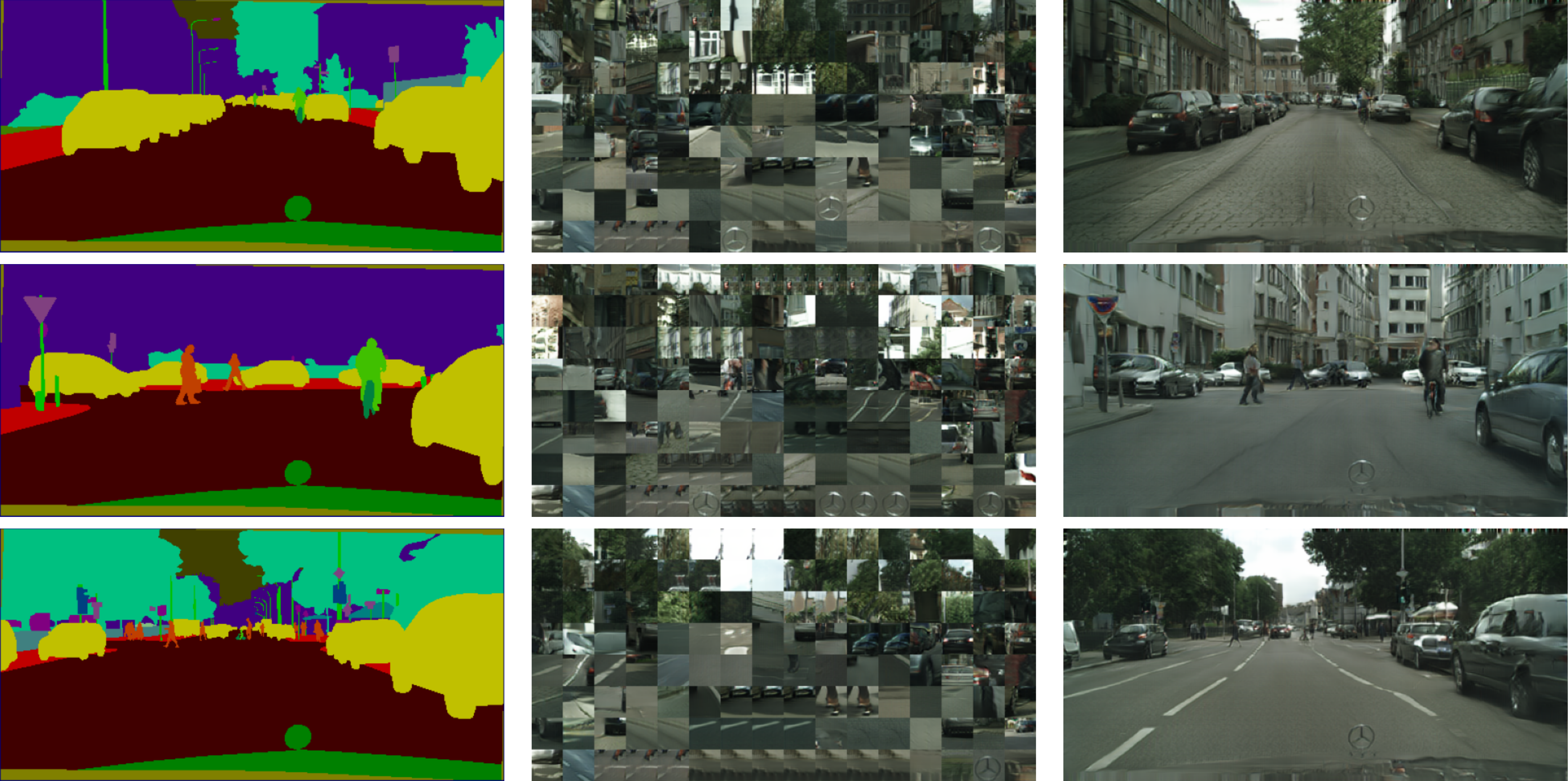}
    \caption{Examples of start images visualizations for the OASIS pipeline. For each feature patch we found the closest one in the bank and then showed corresponding RGB patch of backbone output.}
    \label{fig:sample_oasis}
\end{figure*}

\begin{table*}
  \centering
  \begin{tabular}{c c}
  \textbf{MegaPortraits Predictor} & \\
    \toprule
    Module & (in, out) \\
    \midrule
    Flatten & \\
    Linear + LeakyReLu & (1584, 512)  \\
    Linear + LeakyReLu & (512, 256) \\
    Linear + LeakyReLu & (256, 128) \\
    Linear + LeakyReLu & (128, 64) \\
    Linear + LeakyReLu & (64, 3) \\
    \bottomrule
    Total number of parameters = 1M & FLOPs = 1M 
  \end{tabular}
  \quad
  \begin{tabular}{c c }
  \textbf{OASIS Predictor} &      \\
    \toprule
    Module & (in, out)  \\
    \midrule
    Conv2D + ReLu& (1024, 512)  \\
    ResBlock  & (512, 512) \\
    Flaten & \\
    Linear + ReLu & (10752, 4096) \\
    Linear + ReLu & (4096, 1024) \\
    Linear + ReLu & (1024, 512) \\
    Linear + ReLu & (512, 128) \\
    Linear & (128, 5) \\
    \bottomrule
    Total number of parameters = 58M & FLOPs = 250M 
  \end{tabular}
  \caption{Architecture of the MegaPortraits predictor. Dimensions are in the form (input channels, output channels).}
  \label{tab:avatar_pred}
\end{table*}

\begin{table*}[p]
  \centering
  \begin{tabular}{c c c c c}
 \textbf{OASIS branches} & SF=1/2 & Min. channels = 64 &   &   \\
    \toprule
    Module & Branch 1 & Branch 2 & Branch 3 & Branch 4 \\
    \midrule
    SPADE-ResBlock & (1024, 512, 16, 32) & (1024, 256, 32, 64) & (512, 128, 64, 128) & (256, 64, 128, 256) \\
    SPADE-ResBlock  & (512, 256, 32, 64) & (256, 128, 64, 128) & (128, 64, 128, 256) & (64, 64, 256, 512)  \\
    SPADE-ResBlock  & (256, 128, 64, 128) & (128, 64, 128, 256) & (64, 64, 256, 512) &  \\
    SPADE-ResBlock  & (128, 64, 128, 256) & (64, 64, 256, 512) &  &  \\
    SPADE-ResBlock  & (64, 64, 256, 512) &  &  &  \\
    Conv2D, Tanh    & (64, 3, 256, 512) & (64, 3, 256, 512) & (64, 3, 256, 512) & (64, 3, 256, 512) \\
    \bottomrule
    Total number of parameters  & w/o bank 45.4M & w/ bank 55.7M &
  \end{tabular}
  \\
  \bigskip
    \begin{tabular}{c c c c c}
   \textbf{OASIS branches} & SF=1/3 & Min. channels = 64 &   &   \\
    \toprule
    Module & Branch 1 & Branch 2 & Branch 3 & Branch 4 \\
    \midrule
    SPADE-ResBlock & (1024, 336, 16, 32) & (1024, 168, 32, 64) & (512, 84, 64, 128) & (256, 64, 128, 256) \\
    SPADE-ResBlock  & (336, 168, 32, 64) & (168, 84, 64, 128) & (84, 64, 128, 256) & (64, 64, 256, 512)  \\
    SPADE-ResBlock  & (168, 84, 64, 128) & (84, 64, 128, 256) & (64, 64, 256, 512) &  \\
    SPADE-ResBlock  & (84, 64, 128, 256) & (64, 64, 256, 512) &  &  \\
    SPADE-ResBlock  & (64, 64, 256, 512) &  &  &  \\
    Conv2D, Tanh    & (64, 3, 256, 512) & (64, 3, 256, 512) & (64, 3, 256, 512) & (64, 3, 256, 512) \\
    \bottomrule
    Total number of parameters & w/o bank 35.6M & w/ bank 44.5M & &
  \end{tabular}
  \\
  \bigskip
    \begin{tabular}{c c c c c}
   \textbf{OASIS branches} & SF=1/4 & Min. channels = 64 &   &   \\
    \toprule
    Module & Branch 1 & Branch 2 & Branch 3 & Branch 4 \\
    \midrule
    SPADE-ResBlock & (1024, 256, 16, 32) & (1024, 128, 32, 64) & (512, 64, 64, 128) & (256, 64, 128, 256) \\
    SPADE-ResBlock  & (256, 128, 32, 64) & (128, 64, 64, 128) & (64, 64, 128, 256) & (64, 64, 256, 512)  \\
    SPADE-ResBlock  & (128, 64, 64, 128) & (64, 64, 128, 256) & (64, 64, 256, 512) &  \\
    SPADE-ResBlock  & (64, 64, 128, 256) & (64, 64, 256, 512) &  &  \\
    SPADE-ResBlock  & (64, 64, 256, 512) &  &  &  \\
    Conv2D, Tanh    & (64, 3, 256, 512) & (64, 3, 256, 512) & (64, 3, 256, 512) & (64, 3, 256, 512) \\
    \bottomrule
    Total number of parameters & w/o bank 30.9M & w/ bank 39.1M &
  \end{tabular}
  \caption{Dimensions of modules for all branches in the form of (input channels,~output channels,~image~height,~image~width). In all branches, after each SPADE-ResBlock but the last, we also applied 2D nearest-neighbour upsampling, thus doubling the height and width. When employing the database, the input channels for the first ResBlock in each branch, are multiplied by 1.5.}
  \label{tab:OASIS_m64}
\end{table*}


\begin{table*}
  \centering
  
  \begin{tabular}{c c c c}
   \textbf{MegaPortraits branches} & SF=1/3 & Min. channels = 24 &     \\
    \toprule
    Module & Branch 1 & Branch 2 & Branch 3  \\
    \midrule
    ResBlock2D & (512, 170) & (512, 170) & (512, 170) \\
    ResBlock2D & (170, 170) & (170, 170) & (170, 85)  \\
    ResBlock2D  & (170, 170)& (170, 85) & (85, 42)  \\
    ResBlock2D  & (170, 170)& (85, 42) & \\
    ResBlock2D  & (170, 85) & (42, 24) &   \\
    ResBlock2D  & (85, 42) &  &   \\
    ResBlock2D  & (42, 24) &  &   \\
    ReLu, Conv2D, Tanh & (24, 3) & (24, 3)  & (24, 3)  \\
    \bottomrule
    Total number of parameters & w/ bank 5.6M &  & 
  \end{tabular}
  \\
  
  \begin{tabular}{c c c c}
   \textbf{MegaPortraits branches} & SF=1/6 & Min. channels = 24 &     \\
    \toprule
    Module & Branch 1 & Branch 2 & Branch 3  \\
    \midrule
    ResBlock2D & (512, 85) & (512, 85) & (512, 85) \\
    ResBlock2D & (85, 85) & (85, 85) & (85, 42)  \\
    ResBlock2D  & (85, 85)& (85, 42) & (42, 24)  \\
    ResBlock2D  & (85, 85)& (42, 24) & \\
    ResBlock2D  & (85, 42) & (24, 24) &   \\
    ResBlock2D  & (42, 24) &  &   \\
    ResBlock2D  & (24, 24) &  &   \\
    ReLu, Conv2D, Tanh & (24, 3) & (24, 3)  & (24, 3)  \\
    \bottomrule
    Total number of parameters & w/ bank 2.3M &  & 
  \end{tabular}
  \\
  \bigskip
  
    \begin{tabular}{c c c c}
   \textbf{MegaPortraits branches} & SF=1/8 & Min. channels = 24 &     \\
    \toprule
    Module & Branch 1 & Branch 2 & Branch 3  \\
    \midrule
    ResBlock2D & (512, 64) & (512, 64) & (512, 64) \\
    ResBlock2D & (64, 64) & (64, 64) & (64, 32)  \\
    ResBlock2D  & (64, 64)& (64, 32) & (32, 24)  \\
    ResBlock2D  & (64, 64)& (32, 24) & \\
    ResBlock2D  & (64, 32) & (24, 24) &   \\
    ResBlock2D  & (32, 24) &  &   \\
    ResBlock2D  & (24, 24) &  &   \\
    ReLu, Conv2D, Tanh & (24, 3) & (24, 3)  & (24, 3)  \\
    \bottomrule
    Total number of parameters & w/ bank 1.6M &  & 
  \end{tabular}
  \\
  

  
    \begin{tabular}{c c c c}
   \textbf{MegaPortraits branches} & SF=1/15 & Min. channels = 24 &     \\
    \toprule
    Module & Branch 1 & Branch 2 & Branch 3  \\
    \midrule
    ResBlock2D & (512, 34) & (512, 34) & (512, 34) \\
    ResBlock2D & (34, 34) & (34, 34) & (34, 24)  \\
    ResBlock2D  & (34, 34)& (34, 24) & (24, 24)  \\
    ResBlock2D  & (34, 34) & (24, 24) & \\
    ResBlock2D  & (34, 24) & (24, 24) &   \\
    ResBlock2D  & (24, 24) &  &   \\
    ResBlock2D  & (24, 24) &  &   \\
    ReLu, Conv2D, Tanh    & (24, 3) & (24, 3)  & (24, 3)  \\
    \bottomrule
    Total number of parameters & w/ bank 0.8M &  & 
  \end{tabular}
  \caption{MegaPortraits pipeline. Dimensions of modules for all branches in the form of (input channels,~output channels). The ResBlock2D are made of layers BatchNorm2D, h-swish, Conv2D, BatchNorm2D, h-swish, Conv2D, Conv2D with skipped connections.
  In all branches, before every ResBlock2D, we also applied 2D bilinear upsampling. When employing the database, all input channel numbers must be increased by 3.}
  \label{tab:avatars_ch18}
\end{table*}

\end{document}